\documentclass[lettersize,journal]{IEEEtran}
\usepackage[caption=false,font=normalsize,labelfont=sf,textfont=sf]{subfig}
\usepackage{stfloats}
\usepackage{verbatim}
\usepackage{cite}
\usepackage{amsmath,amssymb,amsfonts}
\usepackage{graphicx}
\usepackage{algpseudocode}
\usepackage{threeparttable}
\usepackage{graphics}
\usepackage{epsfig}
\usepackage{textcomp}
\usepackage{xcolor}
\usepackage{booktabs}
\usepackage[linesnumbered,ruled,vlined]{algorithm2e}
\usepackage{multirow}
\usepackage{makecell}
\usepackage{hyperref}
\usepackage{url}
\usepackage{mathtools}
 \usepackage{bm} 
 \usepackage{pifont}       
\usepackage{bbding}       
\usepackage{fontawesome}  
\definecolor{light-blue}{HTML}{0095d9}
\definecolor{darker-blue}{HTML}{BAC8D3}
\definecolor{dark-green}{HTML}{006400}
\definecolor{dark-blue}{HTML}{1976D2}
\definecolor{dark-purple}{HTML}{8d3fb5}
\definecolor{dark-red}{HTML}{D63C3C}
\usepackage{colortbl}
\hypersetup{
    colorlinks=true,
    linkcolor=dark-red,
    filecolor=magenta,      
    urlcolor=dark-red,
    citecolor=dark-blue,
    pdfpagemode=FullScreen,
}
\newcommand{\cmark}{\ding{51}}
\newcommand{\xmark}{\ding{55}}

\newcommand{\ra}[1]{\textcolor{black}{#1}}
\newcommand{\rb}[1]{\textcolor{black}{#1}}
\newcommand{\rc}[1]{\textcolor{black}{#1}}
\newcommand{\rall}[1]{\textcolor{black}{#1}}

\newcommand*{\tikzmkc}[1]{\tikz[remember picture,overlay,] \node (#1) {};\ignorespaces}
\newcommand{\boxitc}[1]{\tikz[remember picture,overlay]{\node[yshift=3pt,fill=#1,opacity=.1,fit={($(A)+(0.005\linewidth,0.2\baselineskip)$)($(B)+(0.02\linewidth,-0.2\baselineskip)$)}] {};}\ignorespaces}
\SetKwComment{Comment}{$\triangleright$\ }{}
\SetKwComment{CommentL}{$\/\/$}{}
\SetKwFunction{FMain}{Main}
\usepackage{tikz}
\usetikzlibrary{fit,calc}
\definecolor{dark-green}{HTML}{007b43} 
\begin{document}

\title{P$^2$-ViT: Power-of-Two Post-Training Quantization and Acceleration for Fully Quantized Vision Transformer}

\author{Huihong Shi, Xin Cheng, Wendong Mao, and Zhongfeng Wang,~\IEEEmembership{Fellow,~IEEE}

\thanks{This work was supported in part by the National Key R\&D Program of China under Grant 2022YFB4400604 and in part by the Shenzhen Science and Technology
Program 2023A007.}
\thanks{{Huihong Shi and Xin Cheng are with the School of Electronic Science and Engineering, Nanjing University, Nanjing, China ({e-mail:} \{shihh, chengx\}@smail.nju.edu.cn; Wendong Mao is with the School of Integrated Circuits, Sun Yat-sen University, Shenzhen, China (e-mail: maowd@mail.sysu.edu.cn); Zhongfeng Wang is with the School of Electronic Science and Engineering, Nanjing University, and the School of Integrated Circuits, Sun Yat-sen University (email: zfwang@nju.edu.cn).}}
\thanks{Correspondence are addressed to Wendong Mao and Zhongfeng Wang.}}



\maketitle
\begin{abstract}
Vision Transformers (ViTs) have excelled in computer vision tasks but are memory-consuming and computation-intensive, challenging their deployment on resource-constrained devices.
To tackle this limitation, prior works have explored ViT-tailored quantization algorithms but retained floating-point scaling factors, which yield non-negligible re-quantization overhead, limiting ViTs' hardware efficiency and motivating more hardware-friendly solutions. 
To this end, we propose \emph{P$^2$-ViT}, the first \underline{P}ower-of-Two (PoT) \underline{p}ost-training quantization and acceleration framework to accelerate fully quantized ViTs.
Specifically, {as for quantization,} we explore a dedicated quantization scheme to effectively quantize ViTs with PoT scaling factors, thus minimizing the re-quantization overhead. Furthermore, we propose coarse-to-fine automatic mixed-precision quantization to enable better accuracy-efficiency trade-offs.
{In terms of hardware,} we develop {a dedicated chunk-based accelerator} featuring multiple tailored sub-processors to individually handle ViTs' different types of operations, alleviating reconfigurable overhead. Additionally, we design {a tailored row-stationary dataflow} to seize the pipeline processing opportunity introduced by our PoT scaling factors, thereby enhancing throughput.
Extensive experiments consistently validate P$^2$-ViT's effectiveness. {Particularly, we offer comparable or even superior quantization performance with PoT scaling factors when compared to the counterpart with floating-point scaling factors. Besides, we achieve up to $\mathbf{10.1\times}$ speedup and $\mathbf{36.8\times}$ energy saving over GPU's Turing Tensor Cores, and up to $\mathbf{1.84\times}$ higher computation utilization efficiency against SOTA quantization-based ViT accelerators.
Codes are available at \url{https://github.com/shihuihong214/P2-ViT}.}  

\end{abstract}


\vspace{-1em}
\section{{Introduction}}
\IEEEPARstart{T}{hanks} to the powerful capability of Transformers' self-attention mechanism in extracting global information, Vision Transformers (ViTs) have shown great potential and achieved remarkable success in various computer vision (CV) tasks \cite{vit, deit, Graham2021LeViTAV}.
Despite their promising performance, ViTs typically have more parameters and intensive computations than their convolution-based counterparts, yielding higher memory footprint and energy costs during inference. For example, ViT-Large \cite{vit} involves $307$M parameters and $190.7$G FLOPs for achieving $87.76\%$ top-1 accuracy on ImageNet \cite{Deng2009ImageNetAL}.
This challenges the deployment of ViTs on resource-constrained edge devices, calling for effective model compression solutions.

\begin{figure}[t]
	\centerline{\includegraphics[width=0.9\linewidth]{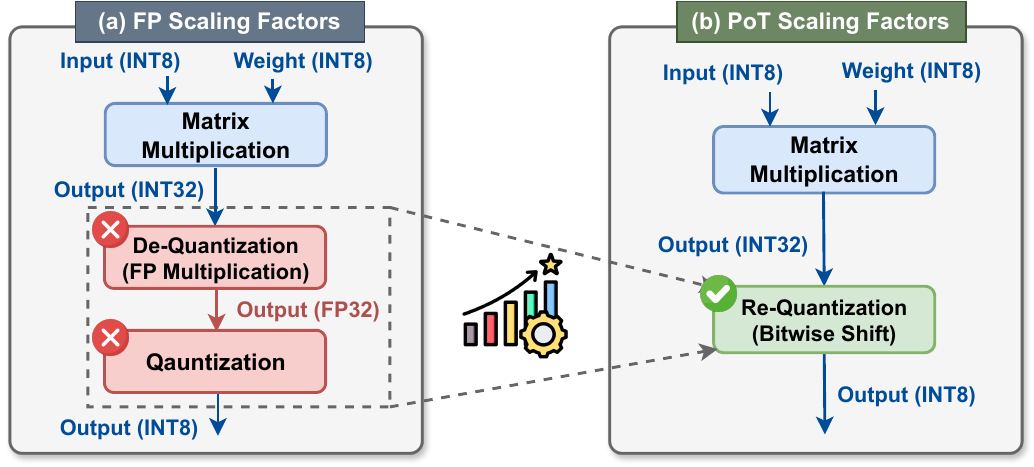}}
	\vspace{-1em}
	\caption{The re-quantization processes with (a) vanilla floating-point (FP) scaling factors and (b) our proposed Power-of-Two (PoT) scaling factors. } 
	\label{fig:re-qaunt} \vspace{-1.2em}
\end{figure}

Model quantization, which converts weights and/or activations from floating-point ones to low-precision integers without modifying network architectures, stands out as a generic and effective model compression technology.
Thus, to facilitate ViTs' deployment, it is natural to adopt quantization to reduce both their memory usage and computational costs.
Specifically, \cite{Liu2021PostTrainingQF} and NoisyQuant \cite{liu2023noisyquant} consider ViTs' specific algorithmic properties to quantize ViTs' linear operations (i.e., matrix multiplications);
FQ-ViT \cite{Lin2021FQViTPQ} introduces several tailored techniques to quantize ViTs' non-linear operations (i.e., LayerNorm and Softmax), thus offering {fully} quantized ViTs.
However, despite the effectiveness of existing ViT quantization methods, they generally target quantization of linear \cite{Yuan2021PTQ4ViTPQ,Liu2021PostTrainingQF, liu2023noisyquant}/non-linear \cite{Lin2021FQViTPQ,huang2023integer}) operations while overlooking and retraining scaling factors in floating-point. This yields non-negligible re-quantization overheads and hinders ViTs' integer-only inference (see Fig. \ref{fig:re-qaunt}a), limiting their speedup on existing hardware platforms \cite{Yao2022RAPQRA, Li2022IViTIQ} and motivating the exploration for more hardware-efficient solutions.

In parallel, prior works \cite{You2022ViTCoDVT, Dass2022ViTALiTyUL, Sun2022VAQFFA, Li2022AutoViTAccAF, dong2023heatvit, huang2023integer} have developed dedicated accelerators to boost ViTs' hardware efficiency from the hardware perspective.
For example, 
{ViTCoD \cite{You2022ViTCoDVT} first leverages pruning to reduce ViTs' redundancy, then develops a dedicated accelerator to gain hardware speedup.}
Additionally, VAQF \cite{Sun2022VAQFFA} and Auto-ViT-Acc \cite{Li2022AutoViTAccAF} implement the acceleration of ViTs on FPGAs with FPGA-aware automatic quantization.
While dedicated accelerators can significantly enhance hardware efficiency, existing ViT accelerators \cite{Sun2022VAQFFA, Li2022AutoViTAccAF, dong2023heatvit} typically focus on the acceleration of matrix multiplications while ignoring the remaining non-linear operations and energy-consuming re-quantization process involved in ViTs. 
These oversights hinder the acceleration potential through techniques like layer fusion and pipeline processing \cite{Alwani2016FusedlayerCA}, thus limiting ViTs' achievable hardware efficiency and real-world applications.
Consequently, it is desirable to build dedicated accelerators to overcome these limitations, i.e., to accelerate fully quantized ViTs alongside minimizing re-quantization overhead to embrace the layer fusion opportunity, which yet is still under-explored.

\begin{figure}[t]
	\centerline{\includegraphics[width=0.9\linewidth]{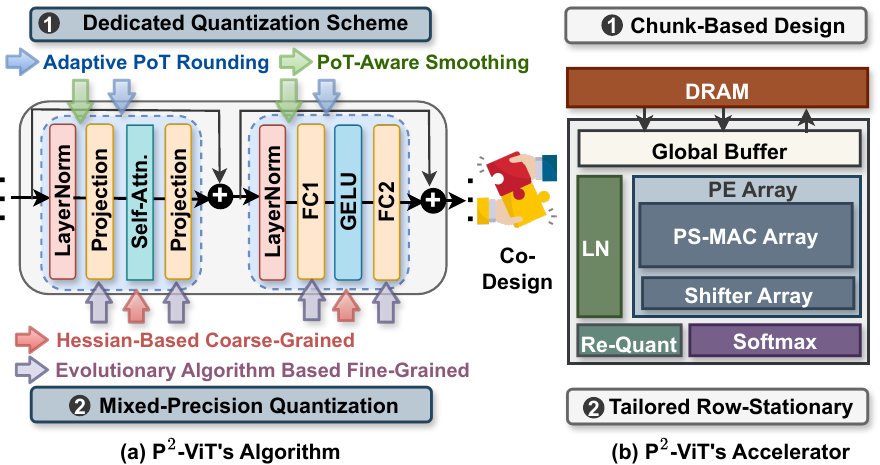}}
	\vspace{-0.6em}
	\caption{The overview of our P$^2$-ViT algorithm and hardware co-design framework. Specifically, (a) P$^2$-ViT's algorithm integrates a dedicated quantization scheme to obtain fully quantized ViTs with Power-of-Two (PoT) scaling factors, and further comprises coarse-to-fine automatic mixed-precision quantization to achieve better accuracy-efficiency trade-offs.
    (b) Furthermore, P$^2$-ViT's dedicated accelerator advocates a chunk-based design incorporating a tailored
    row-stationary dataflow to boost hardware efficiency.} 
	\label{fig:overall} \vspace{-0.8em}
\end{figure}

To this end, to facilitate real-world deployment of ViTs with maintained accuracy, we make the following contributions:
\begin{itemize}
\item We propose \textbf{P$^2$-ViT} (see Fig. \ref{fig:overall}), a \textbf{\underline{P}}ower-of-Two (PoT) \textbf{\underline{p}}ost-training quantization and acceleration framework to accelerate fully quantized Vision Transformers (ViTs).

\item At the algorithmic level, we conduct a comprehensive analysis of ViTs' properties and develop \textbf{a dedicated quantization scheme}. This scheme incorporates innovative techniques such as adaptive PoT rounding and PoT-Aware smoothing, allowing for the efficient quantization of ViTs with PoT scaling factors. By doing this, computationally expensive floating-point multiplications and divisions within the re-quantization process can be traded with hardware-efficient bitwise shift operations (see Fig. \ref{fig:re-qaunt}).
Furthermore, we introduce a \textbf{coarse-to-fine automatic mixed-precision quantization} methodology for better accuracy-efficiency trade-offs.

\item At the hardware level, we build a dedicated accelerator engine to better leverage our algorithmic properties for enhancing hardware efficiency. Specifically, we advocate \textbf{a chunk-based design} with multiple tailored sub-processors (chunks) to separately handle ViTs' different types of operations, 
thus alleviating the reconfigurable overhead.
On top of that, we further propose \textbf{a tailored row-stationary dataflow} to capitalize on the inter- and intra-layer pipeline opportunity introduced by our PoT scaling factors for promoting the throughput.

\item Extensive experiments and ablation studies on various ViT models consistently validate the benefits of our P$^2$-ViT framework in enhancing hardware efficiency while maintaining accuracy. 
\end{itemize}

\textcolor{black}{The rest of the paper is organized as follows}: we first introduce prior efforts and motivations in Sec. \ref{sec:related_work}, then illustrate our quantization algorithm and dedicated accelerator in Sec. \ref{sec:alg} and Sec. \ref{sec:hw}, respectively; Then, Sec. \ref{sec:exp} demonstrates the superiority of our P$^2$-ViT framework via extensive experiments and ablation studies; 
Finally, Sec. \ref{sec:conclusion} summarizes this paper. 
\section{Related Work and Motivations}
\label{sec:related_work}
\subsection{Vision Transformers (ViTs)}
Motivated by the powerful capability of the self-attention mechanism in extracting global information, there has been a growing interest in developing Transformers for CV tasks \cite{vit, deit, Graham2021LeViTAV, Cai2022EfficientViTEL}. 
Among them, ViT \cite{vit} for the first time adopts a pure Transformer model to process sequences of image patches, achieving competitive results over SOTA CNNs when pretrained on extremely large datasets.
On top of that, DeiT \cite{deit} further refines training recipes for ViT and offers comparable results when only pretrained on ImageNet \cite{Deng2009ImageNetAL}, thus saving training costs.
Furthermore, many efforts \cite{Graham2021LeViTAV, Cai2022EfficientViTEL} have been made to boost ViTs' inference efficiency by constructing hybrid architectures to marry the benefits of both CNNs and Transformers. For example, LeViT \cite{Graham2021LeViTAV} replaces the vanilla Transformer structure with a pyramid-like architecture and further introduces an attention bias to integrate positional information.
Different from the above works, which mainly focus on the exploration of either effective training strategies or efficient ViT structures, our P$^2$-ViT targets ViT's quantization and acceleration via algorithm and hardware co-design. 

\vspace{-1em}
\subsection{Model Quantization}
Quantization is a generic and promising model compression solution to reduce both memory footprint and computational costs.
\rc{Current quantization can be divided into two categories: quantization-aware training (QAT) and post-training quantization (PTQ).
The former \cite{Li2022IViTIQ, Jacob2017QuantizationAT, Yao2020HAWQV3DN, Dong2019HAWQV2HA} involves resource-consuming weight fine-tuning to facilitate quantization, offering higher accuracy and lower bits. 
To eliminate this fine-tuning phase and enable rapid deployment, PTQ  \cite{Li2021BRECQPT, adaround, Xiao2022SmoothQuantAA, wei2023outlier} has gained increasing attention.}
Particularly, motivated by ViTs' promising performance in CV tasks, their dedicated PTQ algorithms \cite{Liu2021PostTrainingQF, Yuan2021PTQ4ViTPQ, Lin2021FQViTPQ, liu2023noisyquant, li2023repq} have gained growing attention.
For instance, 
\cite{Liu2021PostTrainingQF} integrates a novel ranking loss into the vanilla quantization loss for preserving the functionality of the self-attention mechanism.
RepQ-ViT \cite{li2023repq} proposes quantization scale reparameterization to enhance quantization accuracy.
Moreover, FQ-ViT \cite{Lin2021FQViTPQ} proposes Power-of-Two Factor (PTF) and Log-Int-Softmax (LIS) to quantize ViTs' non-linear LayerNorm and Softmax, respectively, offering fully quantized ViTs.

\textbf{Motivations.} 
\ra{\underline{(i)} However, as shown in Fig. \ref{fig:re-qaunt}a, scaling factors within the above works are still represented in floating-point, yielding costly floating-point multiplications/divisions for re-quantization and impeding ViTs' integer-only arithmetic during inference \cite{Yao2022RAPQRA, Li2022IViTIQ}.
{\underline{(ii)} To solve this limitation, some works \cite{Li2022IViTIQ, Jacob2017QuantizationAT} tend to implement the floating-point re-quantization via dyadic numbers (DN), thus enabling the execution of ViTs within the integer domain. However, they need costly fine-tuning to facilitate quantization and also involve costly high-bit multiplications within the re-quantization process.
\underline{(iii)} To further enhance re-quantization efficiency, several works \cite{Jain2019TrainedQT, Yao2022RAPQRA} have explored quantization with PoT scaling factors, thus trading the costly floating-point multiplications and divisions
within the re-quantization process with hardware-efficient bitwise shift, as shown in Fig. \ref{fig:re-qaunt}b. However, despite their effectiveness, they are dedicated to CNNs and not directly applicable to ViTs due to ViTs' unique model architecture and specific algorithmic characteristics.}}

\ra{Thus, to boost ViTs' re-quantization efficiency and facilitate their real-world applications, we conduct a comprehensive analysis of ViTs’ properties and develop a dedicated quantization scheme to fully quantize ViTs with PoT scaling factors without any fine-tuning.}

\vspace{-1em}
\subsection{Transformer Accelerators}
Apart from the algorithmic optimization, many works \cite{Lu2021SangerAC, Lu2020HardwareAF, Dass2022ViTALiTyUL, You2022ViTCoDVT, Sun2022VAQFFA, Li2022AutoViTAccAF} have constructed dedicated accelerators to boost Transformers' hardware efficiency.
For example, for NLP tasks, Sanger \cite{Lu2021SangerAC} and DOTA \cite{Qu2022DOTADA} dynamically prune the computation-intensive attention maps in Transformers and further develop a reconfigurable architecture to support the resulting sparsity patterns.
For CV tasks, {ViTCoD \cite{You2022ViTCoDVT} and HeatViT \cite{dong2023heatvit} apply static attention map pruning and adaptive token pruning for ViTs, respectively, then build dedicated accelerators to accelerate resultant sparsity workloads.}
Additionally, VAQF \cite{Sun2022VAQFFA} and Auto-ViT-Acc \cite{Li2022AutoViTAccAF} implement ViTs' acceleration on FPGAs with a hardware-aware automatic quantization precision allocation strategy and mixed-scheme quantization scheme, respectively.

\textbf{Motivations.} 
\ra{The primary {limitation} of existing accelerators tailored for ViT quantization lies in their main focus on accelerating matrix multiplications \cite{Li2022AutoViTAccAF, Sun2022VAQFFA}, while neglecting the energy-consuming re-quantization process.
This oversight restricts the acceleration opportunity provided by strategies such as layer fusion and pipeline processing \cite{Alwani2016FusedlayerCA}.}

\ra{To solve this limitation, we develop a dedicated accelerator to fully unleash our algorithmic benefits, i.e., the offered fully quantized ViTs with PoT scaling factors. Specifically, our accelerator \underline{(i)} supports the acceleration of both linear and non-linear operations within ViTs, and \underline{(ii)} enables efficient on-chip re-quantization processing via bitwise shifts, thus facilitating pipeline processing.
}
\section{P$^2$-ViT's Algorithm}
\label{sec:alg}
In this section, we first introduce preliminaries of ViTs, then illustrate limitations of prior works and offered opportunities in Sec. \ref{sec:limit_opp}. Next, we elaborate P$^2$-ViT's dedicated quantization scheme and coarse-to-fine automatic mixed-precision quantization in Sec. \ref{sec:quantiozation_scheme} and Sec. \ref{sec:mp-quant}, respectively.
\vspace{-1em}

\subsection{Preliminaries}
\label{sec:preliminaries}
\begin{figure}[t]
	\centerline{\includegraphics[width=0.9\linewidth]{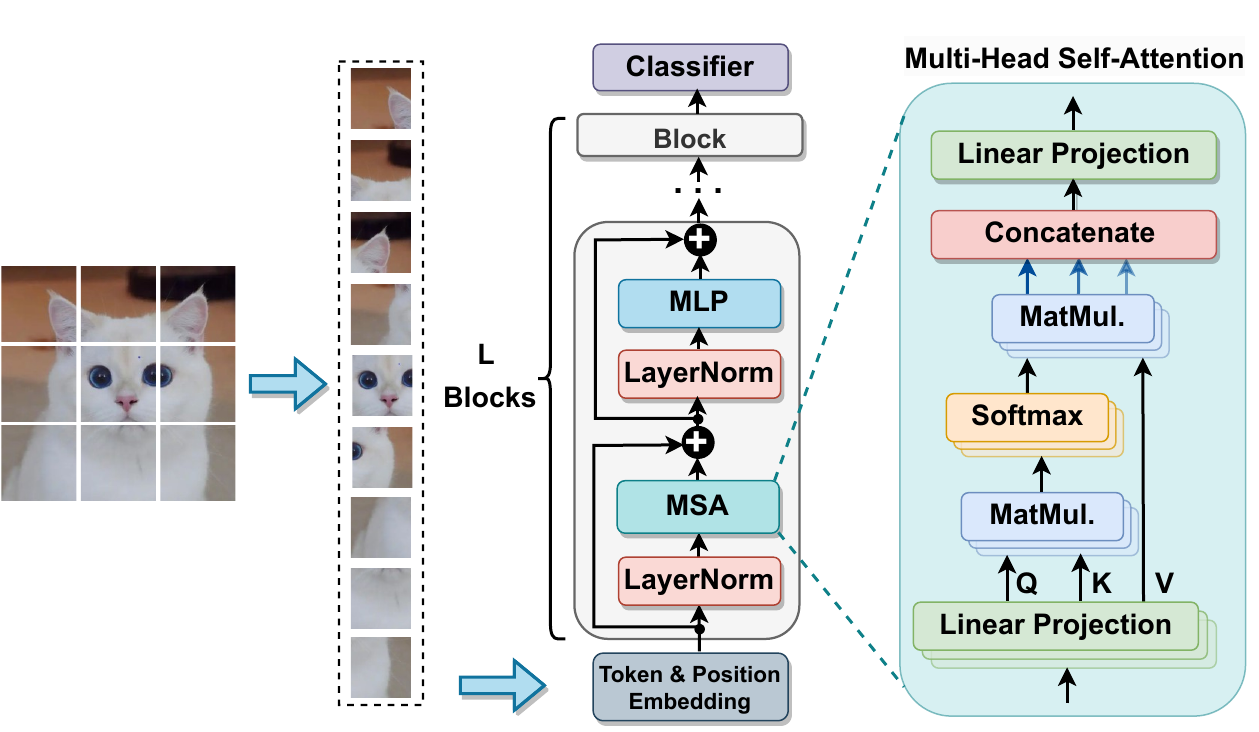}}
	\vspace{-0.6em}
	\caption{The illustration of standard Vision Transformers' (ViTs') architecture (e.g., ViT \cite{vit} and DeiT \cite{deit}) that consists of multiple Transformer blocks. Each block includes a Multi-head Self-Attention module (MSA) and a Multi-Layer Perceptron (MLP). 'MatMul.' is the abbreviation of matrix multiplications. } 
	\label{fig:vit_arch} \vspace{-0.8em}
\end{figure} 

\textbf{Architecture of ViTs.} 
As illustrated in Fig. \ref{fig:vit_arch}, input images are first split into multiple fixed-size patches, then linearly embedded and combined with position embedding to serve as input tokens for Transformer blocks in ViTs. Among them, each block consists of a Multi-head Self-Attention module (MSA) and a Multi-Layer Perceptron (MLP), both of which have LayerNorm (LN) applied before and residual connections added after.
Specifically, \underline{the MSA}, which is the core component of Transformers for extracting global information with boosted performance, first linearly projects input tokens ${X}$ with weights ${W^Q_i}$, ${W^K_i}$, and ${W^V_i}$ in the $i^{th}$ head to generate corresponding queries ${Q_i}$, keys ${K_i}$, and values ${V_i}$, following Eq. (\ref{eq:qkv_proj}). Then, as formulated in Eq. (\ref{eq:attn}), ${Q_i}$ is first multiplied with ${K^{T}_i}$ and then normalized by Softmax (where $d_i$ is the feature dimension of each head) to generate the attention map, which is further multiplied with $V_i$ to obtain the attention output ${A_i}$ of head $i$. Finally, attention outputs from all ${H}$ heads are concatenated and then projected with weights ${W^{O}}$ to produce the final output of MSA (i.e., ${O_\text{MSA}}$ in Eq. (\ref{eq:attn_concat})).
As for \underline{the MLP}, it contains two fully connected layers with GELU between them. 
\begin{equation}
    {[Q_i, K_i, V_i]=X\cdot [W^Q_i, W^K_i, W^V_i]}, \label{eq:qkv_proj}
\end{equation} \vspace{-0.8em}
\begin{equation}
    {A_i}={\text{Softmax}}(\frac{{Q_iK^T_i}}{\sqrt{d_i}})\cdot {V_i}, \label{eq:attn}
\end{equation} \vspace{-0.8em}
\begin{equation}
    {O_\text{MSA}}=\text{concat}({A_0,A_1,...,A_H})\cdot {W^{O}}. \label{eq:attn_concat}
\end{equation} 

\textbf{Full Quantization for ViTs.} 
To obtain fully quantized ViTs, FQ-ViT \cite{Lin2021FQViTPQ} proposes Power-of-Two Factor (\textbf{PTF}) and Log-Int-Softmax (\textbf{LIS}) to quantize the most sensitive non-linear LN and Softmax in ViTs, respectively. 

\begin{figure}[t]
	\centerline{\includegraphics[width=\linewidth]{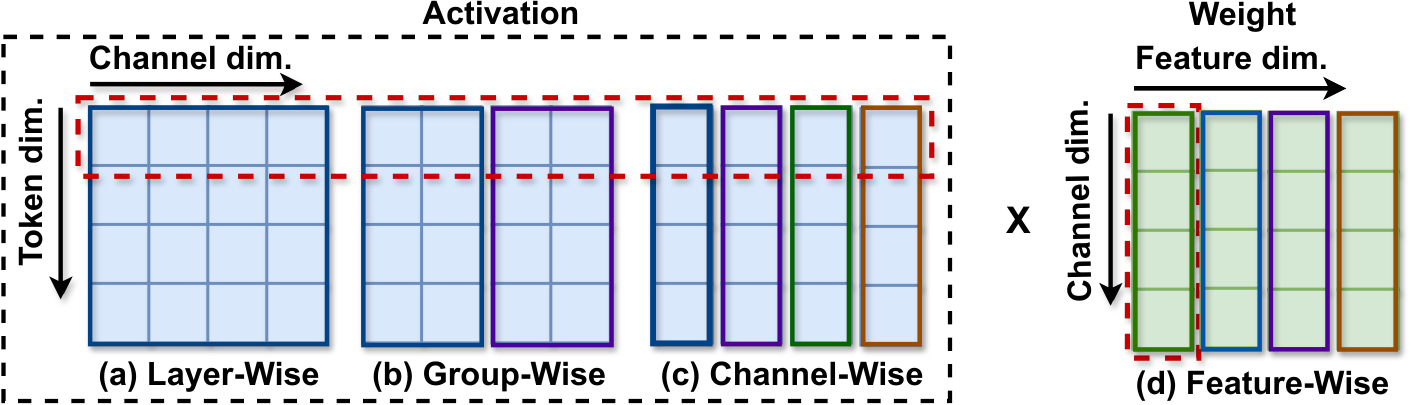}}
	\vspace{-0.6em}
	\caption{Illustrating the (a) layer-wise, (b) group-wise, and (c) channel-wise quantization for activations, and (d) feature-wise quantization for weights.} 
	\label{fig:quantization_schem} \vspace{-1.2em}
\end{figure} 

\begin{figure*}[t]
\centerline{\includegraphics[width=0.98\linewidth]{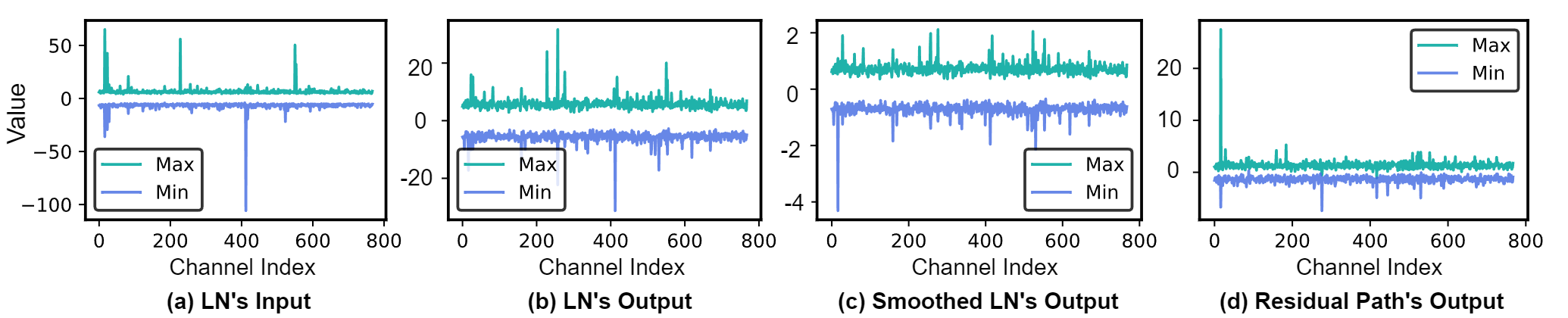}}
	\vspace{-0.6em}
	\caption{The minimum and maximum values of the last (a) LN's input, (b) LN's output, (c) smoothed LN's output, and (d) residual path's output, along the channel dimension in the full-precision ViT-Base, where inputs are randomly sampled from ImageNet \cite{Deng2009ImageNetAL}.} 
	\label{fig:act_distribution} \vspace{-1.2em}
\end{figure*}

\emph{i) PTF.} Due to the extreme inter-channel variations in LN's input activations ({see Fig. \ref{fig:act_distribution}a}), the widely-adopted layer-wise quantization in Fig. \ref{fig:quantization_schem}a, which quantizes the whole layer with the same scaling factor, will result in unbearable quantization errors \cite{Lin2021FQViTPQ, Xiao2022SmoothQuantAA, Shen2019QBERTHB}. 
To enhance quantization accuracy, group-wise quantization \cite{Shen2019QBERTHB} in Fig. \ref{fig:quantization_schem}b or channel-wise quantization \cite{Li2019FullyQN} in Fig. \ref{fig:quantization_schem}c are proposed, which assign group-wise or channel-wise scaling factors to different groups or channels.
However, both of them are hardware unfriendly to LN, where statistics among the channel dimension are required to obtain mean and variance, calling for more efficient solutions.
To this end, FQ-ViT \cite{Lin2021FQViTPQ} proposes \underline{PTF}, which quantizes LN's input ${X}$ via uniform quantization with a global scaling factor $\mathbf{S_g}$ and channel-wise PoT factors $\alpha$ following Eq. (\ref{eq:PTF}) (where $\lfloor \cdot \rceil$ represents rounding to the nearest and $b$ is the quantization bit), thus enabling the calculation of both mean $\mu$ and variance $\sigma^2$ in the integer domain (see the bold part in Eq. (\ref{eq:LN_PTF})).
\begin{equation}
    {X_Q} = Q({X}|b) = \text{clip}(\lfloor\frac{{X}}{2^\alpha \cdot \mathbf{S_g}} \rceil,0,2^b-1). \label{eq:PTF}
\end{equation} \vspace{-0.8em}
\begin{equation}
\begin{aligned}
    \mu({X}) = \mu(2^{\alpha}\mathbf{S_g}\cdot {X_Q})= \mathbf{S_g}\cdot \bm{\mu({X_Q}<<\alpha)}, \\
    \sigma^2({X}) = \sigma^2(2^{\alpha}\mathbf{S_g}\cdot {X_Q})= \mathbf{S_g}\cdot \bm{\sigma^2({X_Q}<<\alpha)}.\label{eq:LN_PTF}
    \end{aligned}
\end{equation} \vspace{-0.5em}

\emph{ii) LIS.} Besides LN's inputs, outliers also exist in ViTs' attention maps ${M}$ (i.e., the outputs of Softmax(Q, K, d) in Eq. (\ref{eq:attn})). 
It motivates FQ-ViT \cite{Lin2021FQViTPQ} to adopt log2 quantization following Eq. (\ref{eq:log2Q}), which is on par with the $8$-bit uniform quantization with $M_Q$ (i.e., the quantized $M$) being represented with only $4$-bit ($b=4$ here).
By doing this, the subsequent matrix multiplications between $M$ and $V$ can be further substituted with hardware-efficient bit-shifts following Eq. (\ref{eq:shift_sv}), where $V_Q$ and $S_v$ are quantized $V$ and its scaling factor,  respectively.
Additionally, FQ-ViT integrates i-exp \cite{Kim2021IBERTIB}, a polynomial approximation of exponential function, with the above log2 quantization to propose \underline{LIS}, thus enabling the calculation of attention outputs (i.e., $A$ in Eq. (\ref{eq:attn})) in the integer domain.
\begin{equation}
    {M_Q} = Q({M}|b) = \text{clip}(\lfloor -\text{log}_{2}{M} \rceil,0,2^b-1), \label{eq:log2Q}
\end{equation} \vspace{-0.8em}
\begin{equation}
    {M}\cdot V = 2^{-{M_Q}}\cdot S_v V_Q=S_v\cdot(V_Q >> M_Q). \label{eq:shift_sv}
\end{equation} \vspace{-0.8em}

\vspace{-2em}
\subsection{Limitations and Opportunities}
\label{sec:limit_opp}
\textbf{Limitations.}
Despite FQ-ViT's effectiveness in offering fully quantized ViTs, their scaling factors are still represented in floating-point, which yields costly floating-point multiplications/divisions for re-quantization, thus limiting ViTs' real-world deployment and calling for more efficient solutions. The re-quantization is expressed in the following bold part:
\vspace{-0.3em}
\begin{equation}
   Y_Q = \frac{Y}{S_y} = \frac{\mathbf{S_xS_w}}{\mathbf{S_y}}\cdot X_Q W_Q,
   \label{eq:re-qaunt}
\end{equation} 
where $Y$ \sloppy is the full-precision output, $Y_Q$, $X_Q$, and $W_Q$ denote quantized output, input, and weight, respectively, and $S_y$, $S_x$, and $S_w$ denote corresponding floating-point scaling factors. 
While several works have explored quantization with PoT scaling factors for CNNs \cite{Jain2019TrainedQT, Yao2022RAPQRA} or implemented floating-point re-quantization via dyadic numbers \cite{Li2022IViTIQ, Jacob2017QuantizationAT}, they either are not directly applicable to ViTs or need costly fine-tuning and involve high-bit multiplications.

\textbf{Opportunity 1: Dedicated Quantization Scheme.} 
To enhance hardware efficiency, we thoroughly analyze properties of ViTs and propose a {dedicated quantization scheme} in Sec. \ref{sec:quantiozation_scheme}, which adopts adaptive PoT rounding and PoT-aware smoothing to convert floating-point scaling factors to their PoT variants without compromising accuracy. By doing this, we can trade the costly floating-point computations in Eq. (\ref{eq:re-qaunt}) with hardware-efficient bit-shifts for minimizing the re-quantization overhead: 
\begin{equation}
   Y_Q = \frac{\mathbf{2^{\alpha_x}2^{\alpha_w}}}{\mathbf{2^{\alpha_y}}}\cdot X_Q W_Q = X_Q W_Q << (\alpha_x+\alpha_w-\alpha_y),
   \label{eq:re-qaunt-PoT}
\end{equation} 
where $\alpha_x$, $\alpha_w$, and $\alpha_y$ indicate exponents of PoT scaling factors for input, weight, and output, respectively.

\textbf{Opportunity 2: Mixed-Precision Quantization.} Additionally, motivated by the fact that layers in ViTs are not equally sensitive to quantization and thus uniformly assigning the same bit-width to all layers is sub-optimal \cite{Liu2021PostTrainingQF}, we further explore {coarse-to-fine automatic mixed-precision quantization} in Sec. \ref{sec:mp-quant} to automatically quantize different layers with different bit-widths for achieving better accuracy-efficiency trade-offs.

\vspace{-1em}
\subsection{Dedicated Quantization Scheme}
\label{sec:quantiozation_scheme}
To facilitate the real-world deployment of fully quantized ViT, we first \underline{(i)} leverage hardware-friendly \textbf{symmetric} uniform quantization for both activations (except the log2 quantization for attention maps $M$ in Eq. (\ref{eq:log2Q}) following LIS in FQ-ViT) and weights, and further intuitively \underline{(ii)} round floating-point scaling factors $S$ to their \textbf{nearest PoT} ones $\hat{S}$ following: 
\begin{equation}
\begin{aligned}
   S=\frac{2\cdot \text{max}(|X|)}{2^b-1}, \
   \alpha = \lfloor \text{log}_2S \rceil, \ \hat{S}=2^\alpha,
   \label{eq:PoT-nearest}
   \end{aligned}
\end{equation} 
where we take the computation of $\hat{S}$ for activation $X$ as the example and $b$ denotes activation's quantization bit.
Note that we leverage the layer-wise quantization in Fig. \ref{fig:quantization_schem}a for activations and feature-wise quantization in Fig. \ref{fig:quantization_schem}d for weights by default, which is the most widely-adopted quantization setting for Transformers \cite{Xiao2022SmoothQuantAA, Lin2021FQViTPQ, Li2019FullyQN, Liu2021PostTrainingQF}.
An exception is the channel-wise quantization in Fig. \ref{fig:quantization_schem}c for LN's input with a global scaling factor and channel-wise PoT factors following PTF in FQ-ViT. Despite the promising hardware efficiency of the above vanilla solution, it yields significant accuracy drops on various ViTs (see {Sec. \ref{sec:alg_results}} for details). 
Thus, to enhance quantization performance, we propose a dedicated quantization scheme, including innovative \underline{(i)} \textbf{adaptive PoT rounding} that adaptively rounds floating-point scaling factors to PoT ones by directly minimizing \rb{quantization perturbation (i.e., quantization errors)} of \emph{activations}, and \underline{(ii)} \textbf{PoT-aware smoothing} that migrates outliers in sensitive activations to less-sensitive weights for facilitating quantization.

\begin{table}[]
\centering
\caption{Top-1 Accuracy Comparisons of Different Rounding Schemes on Various DeiT \cite{deit} Models and Tested on ImageNet \cite{Deng2009ImageNetAL}} \vspace{-0.6em}
\setlength{\tabcolsep}{0.35em}
\resizebox{0.92\linewidth}{!}{
\begin{tabular}{c|cc|ccc} \hline \hline
\textbf{Rounding} & \textbf{PoT-S$^\dagger$}                         & \textbf{W/A/Attn$^*$}      & \textbf{DeiT-Tiny} & \textbf{DeiT-Small} & \textbf{DeiT-Base} \\ \hline \hline
\rowcolor{darker-blue!50} \textbf{Baseline} & \color{black}{\xmark}                  & {8/8/4}  & 71.78              & 79.35               & 81.37              \\ \hline
\textbf{Nearest}  & \multirow{3}{*}{\color{black}{\cmark}} & \multirow{3}{*}{8/8/4}                       & \textbf{69.80}     & 76.15               & \textbf{80.27}     \\
\textbf{Ceil}     &                                        &                        & 69.82              & \textbf{76.80}      & 79.36              \\
\textbf{Floor}    &                                        &                        & 68.29              & 73.26               & 80.09              \\ \hline \hline
\rowcolor{darker-blue!50} \textbf{Baseline} & \color{black}{\xmark}                  & {4/8/4}   & 65.63              & 76.07               & 79.65              \\ \hline
\textbf{Nearest}  & \multirow{3}{*}{\color{black}{\cmark}} & \multirow{3}{*}{4/8/4}                        & \textbf{63.39}     & \textbf{71.94}      & 78.01              \\
\textbf{Ceil}     &                                        &                        & 62.46              & 71.49               & 76.66              \\
\textbf{Floor}    &                                        &                        & 61.49              & 69.04               & \textbf{78.17} \\ \hline \hline   
\end{tabular}} \label{tab:rounding_scheme} 
\begin{tablenotes}
\footnotesize
		\item[$\ast$] $\dagger$ denotes PoT scaling factors; $*$ represents bit-widths for weights, activations, and attention maps, respectively; 
\end{tablenotes} \vspace{-1.2em}
\end{table}

\textbf{(i) Adaptive PoT Rounding.}
\rb{Typically, there are three prevalent approaches to convert floating-point numbers into PoT ones: rounding-to-nearest ($\lfloor \cdot \rceil$ in Eq. (\ref{eq:PoT-nearest})), rounding-to-ceil $\lceil \cdot \rceil$, and rounding-to-floor $\lfloor \cdot \rfloor$. Among them, \textbf{\textit{rounding-to-nearest}} is the most widely adopted method due to its effectiveness in minimizing rounding errors \cite{Elhoushi2019DeepShiftTM}.
However, as revealed in Table \ref{tab:rounding_scheme}, we
have empirically observed that directly applying the predominant rounding-to-nearest approach to convert our floating-point scaling factors to PoT format cannot guarantee optimal performance and generally yields significant accuracy drops.
A similar observation has also been made in \cite{adaround}, which theoretically and experimentally validates that directly rounding all floating-point \textbf{\emph{weights in convolutions}} to their nearest fixed-point values has a catastrophic effect on post-training quantization for CNNs.}

\rb{Inspired by this, we introduce adaptive PoT rounding for \textbf{\textit{scaling factors in ViTs}}, which employs the $L2$ distance between the full-precision and quantized \textit{{activations}} to more accurately round floating-point scaling factors to their PoT variants.
Formally, we employ the quantization perturbation of input activation $X$ to determine its PoT factor $\alpha_x$, and quantization perturbation of output activation $XW$ to find the PoT factor $\alpha_w$ for weight $W$. }
This can be represented as:
\vspace{-0.6em}
\begin{equation}
\begin{aligned}
   \alpha_x = \mathop{\arg\min}_{\alpha_x\in A_x} ||X-\lfloor \frac{X}{2^{\alpha_x}}\rceil \cdot 2^{\alpha_x}||_2, \\ \alpha_w = \mathop{\arg\min}_{\alpha_w\in A_w} ||XW-\lfloor \frac{XW}{2^{\alpha_w}}\rceil \cdot 2^{\alpha_w}||_2, \ &\text{where}  \\
    \ A \in \{\alpha^f-1, \ \alpha^f, \ \alpha^c, \ \alpha^c + 1\}.&
   \label{eq:PoT-rounding}
   \end{aligned}
\end{equation}
\rb{By doing this, we can adaptively choose $\alpha$ from its floor $\alpha^f$ (i.e., $\lfloor \text{log}_2S \rfloor$) or ceiling $\alpha^c$ (i.e., $\lceil \text{log}_2S \rceil$) candidate, aiming to directly minimize the quantization perturbation of \emph{input/output activations} rather than \emph{scaling factors themselves} for boosting accuracy.}
Moreover, we increasingly expand the search space to [$\alpha^f$-1, $\alpha^c$+1] to enable a more powerful search. 

\rb{For example, if the floating-point scaling factor $S$=$5.2$, then $\alpha^f$=$\lfloor \text{log}_2S \rfloor$=$2$ and $\alpha^c$=$\lceil \text{log}_2S \rceil$=$3$. The search space for $\alpha$ will be expanded from \{$\alpha^f$, $\alpha^c$\}=\{$2, 3$\} to [$\alpha^f$-$1$, $\alpha^c$+$1$]=\{$1, 2, 3, 4$\}. Given the relatively small size of the expanded search space, we can iterate the entire search space to identify the optimal value of $\alpha$ following Eq. (\ref{eq:PoT-rounding})}.

\textbf{Identify Bottlenecks.} Despite the effectiveness of adaptive PoT rounding, there still exist non-negligible quantization performance drops (see Sec. {\ref{sec:alg_results}} for details).
To further close the gaps, we first analyze activation distributions in ViTs, as activations are more quantization-sensitive than weights \cite{Lin2021FQViTPQ, Yuan2021PTQ4ViTPQ}. 
From Fig. {\ref{fig:act_distribution}}, where we take activations in the full-precision ViT-Base \cite{vit} as the example, we identify extreme inter-channel variations in \emph{\textbf{three} types of activations}: \underline{(i)} LN's inputs, \underline{(ii)} LN's outputs, and \underline{(iii)} the outputs of residual paths, i.e., MSAs' and MLPs' outputs. 
However, the PTF in FQ-ViT \cite{Lin2021FQViTPQ} can only handle the outlier issue in LN's inputs (i.e., the \textbf{\emph{\underline{first}}} type) as discussed in Sec. \ref{sec:preliminaries}, calling for effective quantization methods for the last two types of activations. 

\textbf{(ii) PoT-Aware Smoothing.} As for channel-wise outliers in LN's outputs $X$ (i.e., the \emph{\textbf{\underline{second}}} type), the most straightforward solution is to adopt channel-wise quantization. However, it is hardware unfriendly when performing matrix multiplications between $X$ and its subsequent layer's weights, where summations along the channel dimension with different scaling factors are required (see Fig. \ref{fig:quantization_schem}c and \ref{fig:quantization_schem}d).
Inspired by SmoothQuant \cite{Xiao2022SmoothQuantAA}, which migrates channel-wise outliers in activations to weights for easing quantization of large language models, we propose PoT-aware smoothing to handle \emph{LN's outputs} $X$ in ViTs.
Specifically, we smooth $X$ with channel-wise PoT migration factors $2^M$ via a mathematically equivalent transformation:
\begin{equation}
\begin{aligned}
\small
   Y = (\frac{X}{2^M}) \cdot (W\cdot 2^M)=\hat{X}\cdot
   \hat{W}, \\ \mathop{s.t.}\ M_i=\lfloor \text{log}_2{\frac{\text{max}(|X_i|)^\beta}{\text{max}(|W_i|)^{1-\beta}}} \rceil,
   \label{eq:PoT-smoothing}
   \end{aligned}
\end{equation} 
where $W$ is weight of subsequent layer, $i$ is the channel/feature index of $X$/$W$, and $\beta$ is a hyper-parameter controlling the migration strength.
By doing this, we can largely decrease the range of absolute values along the channel dimension in LN's outputs, as verified by Figs. \ref{fig:act_distribution}b and \ref{fig:act_distribution}c.   
After the transformation,
we apply layer-wise and channel-wise quantization with PoT scaling factors (via Eq. (\ref{eq:PoT-rounding})) on top of smoothed activations $\hat{X}$ and transformed weights $\hat{W}$, respectively.
Note that we can pre-compute $\hat{W}$ before deployment to avoid on-chip computation overheads of weights. In contrast, as for activations, which depend on input images during inference and thus cannot be pre-smoothed, we fuse their PoT migration factors $2^M$ with the PoT scaling factor $2^{\alpha_{\hat{x}}}$ of $\hat{X}$ to generate the fused scaling factor $2^{\hat{\alpha}_{x}}$ for preventing the online transformation, which can be defined as:
\begin{equation}
   \hat{X}_Q = \frac{\hat{X}}{2^{\alpha_{\hat{x}}}} = \frac{X}{2^M\cdot 2^{\alpha_{\hat{x}}}} = \frac{X}{2^{\hat{\alpha}_x}},
   \label{eq:PoT-smoothing-fuse_s}
\end{equation} 
where $\hat{X}_Q$ is the quantized $\hat{X}$.
\rb{It is noteworthy that due to the effectiveness of our proposed PoT-ware smoothing in deriving PoT migration factors, the energy-consuming floating-point divisions can be traded with hardware-friendly bitwise shifts here to enhance hardware efficiency.}

As for residual paths' outputs (i.e., the above \emph{\textbf{\underline{third}}} type), which have no direct subsequent layers and thus do not need to be multiplied with weights and summed along the channel dimension, the vanilla channel-wise quantization is enough.

\vspace{-0.6em}
\subsection{Coarse-to-Fine Automatic Mixed-Precision Quantization}
\label{sec:mp-quant}
As different layers play different roles and exhibit different sensitivities, mixed-precision quantization, which assigns higher bit-widths to more sensitive layers for boosting accuracy and lower bit-widths to less sensitive layers for boosting hardware efficiency, is highly desired.
\rb{To avoid intensive human efforts in the bit allocation of mixed-precision quantization,}
\rb{we propose coarse-to-fine automatic mixed-precision quantization, which incorporates the coarse-to-fine thought to marry the benefits of the existing Hessian-based method \cite{Dong2019HAWQV2HA}
and evolutionary search, thus enhancing both efficiency and accuracy.} Specifically, \underline{(i)} we first leverage the \emph{Hessian-based sensitivity measurement paired with Pareto-frontier-based bit-width allocation} \cite{Dong2019HAWQV2HA} to coarsely determine the bit-width of each layer for boosting search efficiency; \underline{(ii)} On top of that, we fine-tune the obtained bit-width configuration via \emph{evolutionary search} to boost accuracy.  
Note that we only apply mixed-precision quantization for \textbf{weights}, as activations are more sensitive to quantization, i.e, converting them to low bits will yield catastrophic accuracy drops in post-training quantization as verified in Table {\ref{table:acc_under_diff_formats}}.

\begin{table}[]
\centering
\caption{Accuracy comparisons between Base PTQ and FQ-ViT \cite{Lin2021FQViTPQ} on DeiT-Tiny under different formats. W8/A8 represents that weights and activations are both quantized to $8$ bits} \vspace{-0.6em}
\resizebox{\linewidth}{!}{
\begin{tabular}{c|cc|cc|c}
\hline \hline  
\textbf{Format} & \textbf{W8/A8} & \textbf{W4/A8} & \textbf{W8/A4} & \textbf{W4/A4} & \textbf{Full Precision}                  \\ \hline \hline
\textbf{Base PTQ$^*$}    & 71.78  & 65.63  & 0.26  & 0.27  & \multirow{2}{*}{72.21} \\ \cline{1-5}
\textbf{FQ-ViT} & 71.07  & 64.86  & 0.10  & 0.10  &                       \\ \hline \hline 
\end{tabular}} 
\begin{tablenotes} \label{table:acc_under_diff_formats}
		\footnotesize
		\item[*] * indicates that both LN and Softmax are not quantized. 
  \end{tablenotes} \vspace{-0.6em}
  \vspace{-0.8em}
\end{table} 

\textbf{(i) Hessian-Based Coarse-Grained Mixed-Precision Quantization.}
To enable Hessian-based sensitivity measurement, we need to compute the Hessian trace for each layer. However, the generation of Hessian matrix $H$ (i.e., the second-order operator) is computationally infeasible as the dimension of each layer in ViTs is quite large. 
To handle this limitation and offer an efficient trace (i.e., the sum of eigenvalues) estimation, we leverage the fast algorithm in \cite{Dong2019HAWQV2HA}, which adopts the matrix-free method \cite{Martens2010DeepLV, yao2018large, yao2018hessian,bai1996some} and Hutchinson algorithm \cite{avron2011randomized} to compute the Hessian trace without explicitly forming $H$.
Specifically, we first compute the Hessian matvec $Hz$ via the matrix-free method:
\begin{equation}
   Hz=\frac{\partial (g^T z)}{\partial W}, \ \text{where} \ g=\frac{\partial L}{\partial W},
   \label{eq:Hessian_matvec}
\end{equation} 
where $g$ is the gradient of loss $L$ with respect to weights $W$, and $z$ denotes a random vector, whose elements are sampled from Gaussian or Rademacher distribution. 
On top of that, the Hutchinson algorithm is applied to enable the fast estimation of the Hessian trace $Tr(H)$:
\begin{equation} \vspace{-0.4em}
  Tr(H) \approx \frac{1}{m}\sum_{i=1}^{m} z_i^T H z_i.
   \label{eq:Hessian_trace}
\end{equation} \vspace{-0.4em}

After obtaining $Tr(H)$, we further leverage the Pareto-frontier-based method to enable the automatic bit-width allocation for ViTs.
Concretely, we sort candidate bit configurations based on their total second-order perturbation:
\begin{equation} \vspace{-0.4em}
  \Omega = \sum_{i=1}^{N}\Omega_i = \sum_{i=1}^{N} \overline{Tr}(H_i)\cdot ||W_i^Q-W_i||_2,
   \label{eq:omega}
\end{equation} 
where $i$ indicates the $i^{th}$ layer of total $N$ layers, $\overline{Tr}(H_i)$ is the averaged Hessian trace of data samples, and $||W_i^Q-W_i||_2$ denotes the $L2$ distance of quantization perturbation of weights. 
Given a target model size, we select bit configurations with minimal $\Omega$. 

\textbf{(ii) Evolutionary Algorithm Based Fine-Grained Mixed-Precision Quantization.}
\rb{To boost quantization accuracy, we further fine-tune obtained bit configurations via the evolutionary algorithm (more specifically, the genetic algorithm), which is extensively utilized for various search and optimization problems \cite{AlphaNet, Shi2024NASAFFS} due to its efficacy, simplicity, and robustness.
Specifically, unlike the traditional {\textit{brute force search}}, which requires the iteration of the entire search space to identify the optimal solution and will become prohibitively time-consuming if the search space is vast, the evolutionary-based method offers better search efficiency.
Additionally, in contrast to {\textit{gradient-based methods}} \cite{FBNet, NASA}, which necessitate converting discrete search spaces into differentiable ones and are sensitive to hyper-parameter selection, evolutionary-based search is more user-friendly and robust.} 

In terms of {evolutionary search process}, we first \underline{(1)} initialize a population of bit configurations via selecting those with minimal $\Omega$ in Eq. (\ref{eq:omega}) under the given model size; then \underline{(2)} conduct \textbf{crossover} (i.e., generate a new bit configuration via randomly selecting two configurations from the population and inserting one configuration’s bit to the other configuration) and \textbf{mutation} (i.e., generate a new bit configuration from a randomly selected configuration via randomly permuting its bit to another choice) several times to enlarge the population. 
\underline{(3)} After that, we update the population by sorting candidates upon quantization accuracy and only keeping top-k ones under the given model size. We next iterate the above steps $2$ and $3$ until reaching a predefined maximal cycle number. 

\vspace{-0.5em}
\subsection{Overall Algorithm Pipeline}
{
\begin{figure}[!t]

\makeatletter
\newcommand{\removelatexerror}{\let\@latex@error\@gobble}
\makeatother
\noindent\hspace{1.5pt}
\begin{minipage}{0.47\textwidth-1.5pt}
\centering

\begingroup
\removelatexerror

\begin{algorithm}[H]

\caption{P$^2$-ViT's Post-Training Quantization}
\label{alg:ptq}

\SetAlgoLined
\KwIn{Full precision ViT and model size constraint} 
\KwOut{Quantized ViT with power-of-two scaling factors and mixed-precision}

\CommentL{\textcolor{dark-red}{{\textit{// Dedicated Quantization Scheme}}}}
Leverage \textit{channel-wise} quantization for weights and \textit{layer-wise} quantization for activations by default

Employ {\textit{PTF}} via Eq. (\ref{eq:PTF}) and {\textit{LIS}} via Eq. (\ref{eq:log2Q}) to quantize {LN's inputs} and {attention maps}, respectively, following FQ-ViT \cite{Lin2021FQViTPQ}



    \tikzmkc{A}
    Adopt proposed \textbf{\textit{PoT-aware smoothing}} following Eq. (\ref{eq:PoT-smoothing}) to quantize {LN's outputs}  

    Use \textit{channel-wise} quantization for residual paths’ outputs \ \ \ \ \ \ \ \ \ \ \ \ \ \ \ \ \ \ \ \ \ \ \ \ \ \ \ \ \ \ \ \  \ \ \ \ \ \ \ \ \ \ \ \ \ \ \ \ \ \ \ \ \ \ \ 
    \tikzmkc{B} \boxitc{dark-red}

    \tikzmkc{A}
    Employ \textit{\textbf{adaptive PoT rounding}} to convert all floating-point scaling factors to PoT via Eq. (\ref{eq:PoT-rounding}) \ \ \ \  
    \tikzmkc{B} \boxitc{dark-red}

    \CommentL{\textcolor{dark-blue}{{\textit{// Coarse-to-Fine Automatic Mixed- Precision Quantization}}}}
    
    \tikzmkc{A}
    Adopt \textit{\textbf{Hessian-based metric}} to coarsely obtain bit configurations under given model size constraint via Eq. (\ref{eq:omega})

     Fine-tune bit configurations via \textbf{\textit{evolutionary search}}  
    \tikzmkc{B} \boxitc{dark-blue}

\end{algorithm}

\endgroup

\end{minipage}
\vspace{-1.5em}
\end{figure}
}
\rall{The overall algorithm pipeline of our post-training quantization is summarized in Alg. \ref{alg:ptq}.
In particular, FQ-ViT \cite{Lin2021FQViTPQ} has observed extreme outliers in LN's inputs and attention maps, and proposes PTF and LIS to effectively handle them. 
On top of that, we further identify substantial
inter-channel variations in LN's outputs and residual paths' outputs, and advocate adopting \textit{\textbf{PoT-aware smoothing}} and \textit{channel-wise quantization} to facilitate their quantization, respectively. After tackling these critical challenges, we introduce \textbf{\textit{adaptive PoT rounding}} to adaptively convert all floating-point scaling factors to PoT format, thus significantly reducing the re-quantization overhead.  
To further enhance quantization efficiency, we propose a coarse-to-fine automatic mixed-precision quantization approach to automatically determine the optimal bit configurations under the given model size constraint. This comprises a \textbf{\textit{Hessian-based coarse-grained}} bit allocation to boost search efficiency, coupled with a \textbf{\textit{fine-grained evolutionary search}} to refine accuracy.
}
\vspace{0.5em}
\section{P$^2$-ViT's Accelerator}
\label{sec:hw}
In this section, we first present design considerations of our P$^2$-ViT's accelerator, then introduce the chunk-based architecture in Sec. \ref{sec:chunk_design} to alleviate the reconfigurable overhead. Finally, Sec. \ref{sec:rs_dataflow} illustrates the tailored row-stationary dataflow proposed to facilitate both inter- and intra-layer pipelines. 

\vspace{-0.6em}
\subsection{Design Considerations}
\label{sec:hardware_considerations}
\textbf{Design Consideration \# 1: Micro-Architecture.}
\rb{To implement the hardware acceleration of fully quantized ViTs and exploit the \textbf{\textit{pipeline opportunity}} introduced by our \textbf{\textit{PoT
scaling factors}}, our P$^2$-ViT's accelerator is expected to support:
\underline{(i)} \emph{linear operations}, i.e., matrix multiplications ({MatMuls}); 
\underline{(ii)} \emph{non-linear operations}, including LN and Softmax;
as well as \underline{(iii)} \emph{re-quantization operations}, where shifters are required to re-quantize output activations with PoT scaling factors.
Typically, two hardware designs can be considered.
The first one is \underline{(i)} the \emph{reconfigurable design} that leverages a single processor to support multiple operation types, where the key point is to minimize the overhead for supporting such reconfigurability.
The second is \underline{(ii)} the \emph{chunk-based design} that advocates multiple tailored sub-processors/computing engines (also dubbed chunks) to separately process different operation types, where the reconfigurable overhead is eliminated at the cost of fewer hardware resources being allocated for each chunk given the area constraint.}

\rb{Given the negligible computational similarity across the above operation types and that MatMuls dominate the computational costs, we consider the \textbf{second design} and assign a larger chunk for costly multiplications while smaller chunks for other operations, including LN, Softmax, and re-quantization operations. Crucially, this kind of design allows different types of operations to be separately executed on dedicated chunks, thus multiple computations can be simultaneously executed to facilitate pipeline processing.}        

\textbf{Design Consideration \# 2: Dataflow.}
There exist both inter-layer and intra-layer data dependencies during the inference of ViTs. 
Specifically, \underline{(i)} as for the inter-layer dependency, the outputs of LN also serve as inputs to the first linear layer in either MSA or MLP, as depicted in Fig. \ref{fig:vit_arch}.
\underline{(ii)} As for the intra-layer dependency, as expressed in Eq. (\ref{eq:attn}), the outputs of MatMuls between $Q$ and $K^T$ serve as inputs for Softmax, and then the normalization outputs need to be further multiplied with $V$.
Considering the property of our adopted chunk-based design as discussed above, i.e., the limited computation resources need to be partitioned into multiple smaller dedicated chunks for separately supporting different types of operations, such data dependency will lead to high latency and low hardware utilization if we \emph{sequentially} process each step.
{Fortunately, our PoT scaling factors can minimize the re-quantization overhead and thus enable pipeline processing, making it possible to translate this challenge into an acceleration opportunity.}
To seize this \underline{\textit{\textbf{opportunity}}}, it is essential to design an appropriate dataflow for our dedicated accelerator to facilitate both inter-layer and intra-layer pipelines, i.e., multiple steps can be \emph{concurrently} executed by their corresponding chunks, thus overcoming latency bottlenecks and boosting throughput.

\begin{figure}[t]
	\centerline{\includegraphics[width=\linewidth]{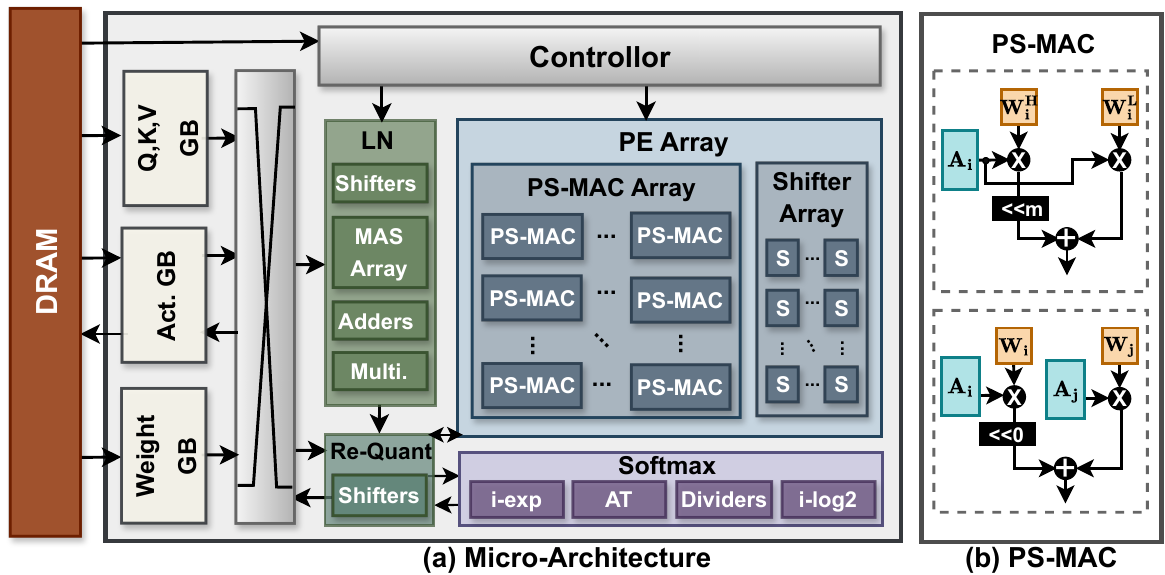}}
	\vspace{-1em}
	\caption{(a) depicts the micro-architecture of P$^2$-ViT's accelerator. The left shows the memory hierarchy to facilitate data reuse and the right gives details of chunks in our chunk-based design for separately processing different operation types in fully quantized ViTs.
    Q, K, V, and GB represent query, key, value, and global buffers, respectively. 
    MAS, PS-MAC, AT, and S denote the multiplication and summation, precision-scalable multiplier and accumulation unit, adder tree, and shifter, respectively.
    (b) illustrates our PS-MAC unit.} 
	\label{fig:hardware_arch} \vspace{-1em}
\end{figure} 

\subsection{Chunk-Based Micro-Architecture}
\label{sec:chunk_design}
As shown in Fig. \ref{fig:hardware_arch}a, to alleviate the reconfigurable overhead of a single processor for simultaneously supporting multiple operation types in fully quantized ViTs, we advocate a chunk-based design. It integrates multiple tailored sub-processors (chunks), such as the Processing Element (PE) array, LN module, Softmax module, and re-quantization module, to separately handle different operations.
Specifically, the \underline{PE array} consists of a Precision-Scalable Multiplier and ACcumulation (PS-MAC) array and a shifter array. The former leverages PS-MACs to conduct MatMuls under different bit-widths for supporting P$^2$-ViT's mixed-precision quantization introduced in Sec. \ref{sec:mp-quant}. The latter aims to fully unleash the benefit of log2 quantization in LIS, i.e., utilize shifters to trade MatMuls between attention outputs $M$ and queries $V$ with bitwise shifts following Eq. (\ref{eq:shift_sv}).
Besides, the \underline{LN module} is equipped with shifters, a Multiplication And Summation (MAS) array, adders, and multipliers to process LN, where input activations are quantized following PTF to boost quantization performance while maintaining hardware efficiency as introduced in Sec. \ref{sec:preliminaries}.
The \underline{Softmax module} is constructed for the LIS introduced in Sec. \ref{sec:preliminaries}, which integrates i-exp \cite{Kim2021IBERTIB} to offer integer-only arithmetic for Softmax and log2 quantization for attention maps to boos quantization performance \cite{Lin2021FQViTPQ}.        
Moreover, a \underline{re-quantization} module featuring shifters is adopted to re-quantize output activations with PoT scaling factors, thus trading the costly floating-point multiplications/divisions in Eq. (\ref{eq:re-qaunt}) with lower-cost bitwise shifts in Eq. (\ref{eq:re-qaunt-PoT}) for boosting the re-quantization efficiency.
Next, we will elaborate on these modules in detail. 

\textbf{PS-MAC.}
To implement our P$^2$-ViT's automatic coarse-to-fine mixed-precision quantization, we construct the PS-MAC. As shown in Fig. \ref{fig:hardware_arch}b, it consists of two low-precision multipliers, as well as a shifter and an adder, to simultaneously support one multiplication between the activation and high-bit weight and two multiplications between activations and low-bit weights.
Specifically, \underline{(i)} as illustrated in Fig. \ref{fig:hardware_arch}b (top) and expressed in Eq. (\ref{eq:ps-mac}), for the first case, the high-bit weight $W_i$ is partitioned into $m$-bit MSBs (Most Significant Bits) $W^H_i$ and the remaining $m$-bit LSBs (Least Significant Bits) $W^L_i$, which are separately passed to two low-bit multipliers to be multiplied with the activation $A_i$. Then, the partial sum of $W^H_i$ is shifted and further added with that of $W^L_i$ to generate the final output.
\underline{(ii)} As shown in Fig. \ref{fig:hardware_arch}b (bottom), for the second case, two different low-bit weights paired with their corresponding activations, i.e., ($W_i$, $A_i$) and ($W_j$, $A_j$), are sent to two multipliers for multiplications, respectively. Then, the two outputs are directly added, thus the parallelism is two.  
\begin{equation}
    W_i\times A_i = (W_i^H\cdot2^m+W_i^L)\times A_i = (W_i^H\times A_i)\cdot2^m + W_i^L\times A_i.
    \label{eq:ps-mac}
\end{equation}

\begin{figure}[t]
\centerline{\includegraphics[width=\linewidth]{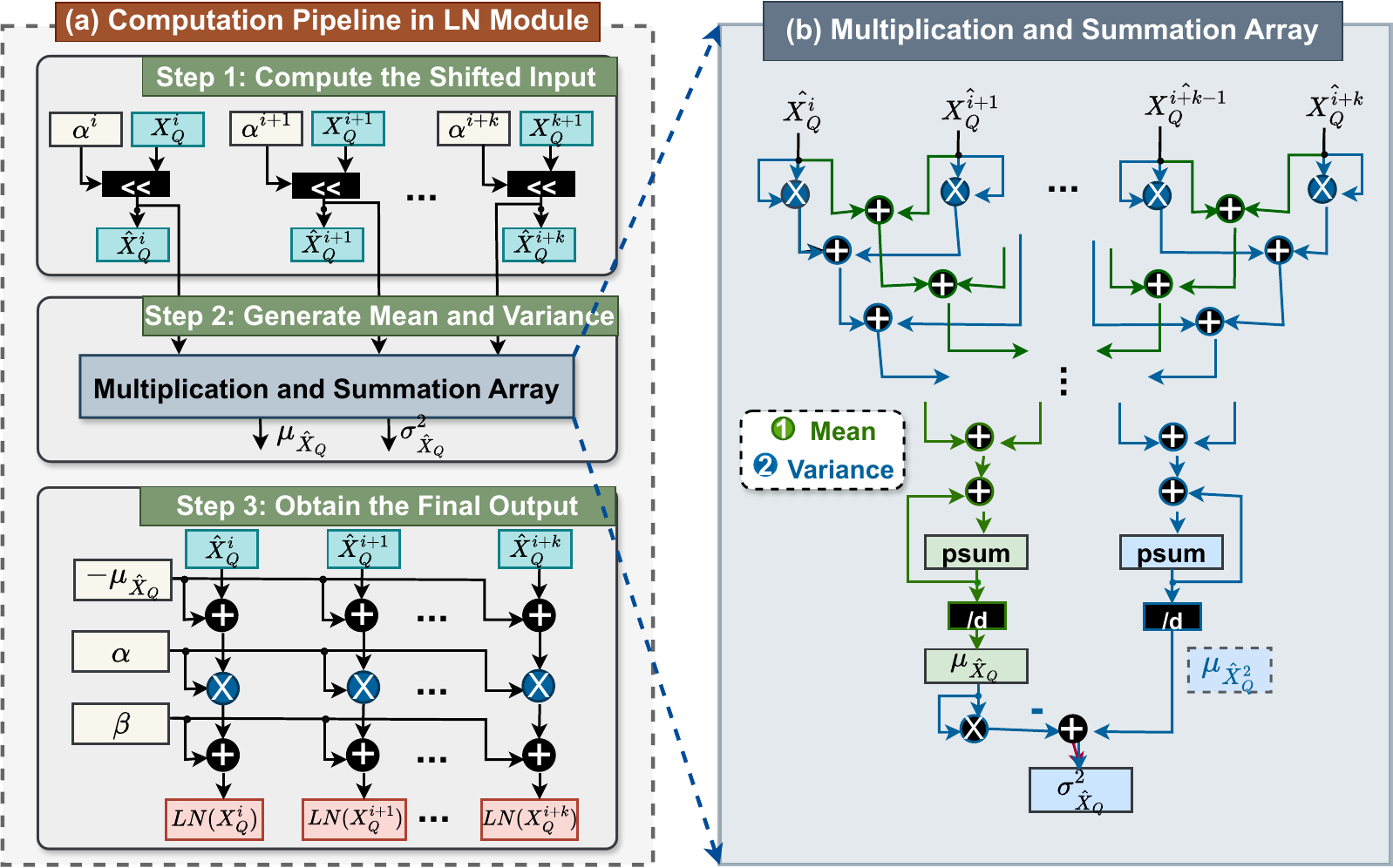}}
	\vspace{-0.6em}
	\caption{The illustration of our LN module, where (a) shows the three-step computation pipeline of the LN processing, and (b) gives details of the multiplication and summation (MAS) array to generate the mean and variance.} 
	\label{fig:LN} \vspace{-1em}
\end{figure} 
\textbf{LN Module.}
To support PTF, which advocates a global scaling factor $S_g$ and channel-wise PoT factors $\alpha$ for LN's inputs following Eq. (\ref{eq:PTF}) to properly handle their extreme inter-channel variations while enabling the computation of mean and variance with integer-only arithmetic, we build the LN module in Fig. \ref{fig:LN}.
As shown in Fig. \ref{fig:LN}a, the computation pipeline of our LN module can be divided into three steps: computing the shifted input via shifters, then generating mean and variance via the MAS array, and finally obtaining LN's output via adders and multipliers.  
Concretely, \underline{(i)} to avoid floating-point computations and provide integer-only arithmetic for calculating mean and variance as expressed in Eq. (\ref{eq:LN_PTF}), we first adopt shifters to shift elements in each row of PTF-quantized input $X_Q$ of LN with their corresponding channel-wise PoT factors $\alpha$, thus obtaining the shifted input $\hat{X}_Q$.

\underline{(ii)} After the above pre-processing, we can directly leverage an adder tree to sum elements in $\hat{X}_Q$ along the channel dimension to compute the corresponding mean $\mu_{\hat{X}_Q}$ following Eq. (\ref{eq:mean}), where $N$ is the number of channels in $\hat{X}_Q$.
As for the generation of variance $\sigma_{\hat{X}_Q}^2$, there are two ways: Eq. (\ref{eq:var_1}) and Eq. (\ref{eq:var_2}).
As for the former one, several extra cycles are needed to pre-compute $\mu_{\hat{X}_Q}$ before we start the accumulation along the channel dimension to obtain variance, yielding delays.
To break this latency bottleneck, we develop the MAS array in Fig. \ref{fig:LN}b to implement the \textbf{latter one} instead, which simultaneously accumulates $\hat{X}_Q$ and $\hat{X}_Q^{2}$ to compute $\mu_{\hat{X}_Q}$ and $\mu_{\hat{X}_Q^2}$, respectively, thus eliminating the latency overhead from the pre-calculation of $\mu_{\hat{X}_Q}$ and enhancing the throughput.  
\begin{equation}
    \mu_{\hat{X}_Q} = \frac{1}{N}\sum_{i=1}^{N}\hat{X}_Q^{i},
    \label{eq:mean}
\end{equation}
\begin{equation}
    \sigma_{\hat{X}_Q}^2 = \frac{1}{N}\sum_{i=1}^{N}[(\hat{X}_Q^{i}-\mu_{\hat{X}_Q})^2]. \label{eq:var_1}
\end{equation}
\begin{equation}
    \sigma_{\hat{X}_Q}^2 
    = \frac{1}{N}\sum_{i=1}^{N} {(\hat{X}_Q^{i})}^2 - (\frac{1}{N}\sum_{i=1}^{N}\hat{X}_Q^{i})^2 
    = \mu_{\hat{X}_Q^2} - \mu_{\hat{X}_Q}^2. \label{eq:var_2}
\end{equation}

\underline{(iii)} After that, 
as shown in Fig. \ref{fig:LN}a (below), we can generate LN's final output via adders and multipliers following:  
\begin{equation}
    LN(X_Q) = \frac{X_Q-\mu_{\hat{X}_Q}}{\sqrt{\sigma_{\hat{X}_Q}^2+\varepsilon}}\cdot \gamma +\beta = (X_Q-\mu_{\hat{X}_Q})\cdot \alpha + \beta, \label{eq:ln}
\end{equation}
where $\gamma$ and $\beta$ are two learnable parameters in LN and $\varepsilon$ is a small constant used to avoid the denominator from being zero. 

\textbf{Softmax Module.}
To achieve LIS in Eq. (\ref{eq:LSI}), where $X_Q$ and $b$ are the quantized input and number of quantization bits, respectively, 
we construct the Softmax module in Fig. \ref{fig:hardware_arch}a (lower right), which contains an i-exp module, an Adder Tree (AT), dividers, and an i-log2 module, to offer integer-only arithmetic for Softmax.
Specifically, the \underline{i-exp module} is constructed to implement the i-exp function \cite{Kim2021IBERTIB, Lin2021FQViTPQ}, which approximates the non-linear exponential function via a second-order polynomial to enable the calculation of Softmax in the integer domain.
Furthermore, the \underline{AT} and \underline{dividers} are utilized to conduct the channel-wise summations and element-wise divisions in Softmax, respectively. 
Additionally, the \underline{i-log2 module} is developed to implement log2 quantization with integer-only arithmetic \cite{Lin2021FQViTPQ}, i.e., it first utilizes shifters to find the index $i$ of the first non-zero bit of input, then adds $i$ with the value of the $(i\mbox{-}1)^{th}$ bit to obtain the result.
For example, if the input of our i-log2 module is ${(0011\ 1001)}_2$, then the index $i$ of the first non-zero bit is $5$ and the value of the $4^{th}$ bit is $1$, thus the log2-quantized input will be $6$.
\begin{equation}
\begin{aligned}
   \text{LIS}(X_Q) &=\text{clip}(\text{log}_2\lfloor \text{Softmax}(X_Q) \rceil,0,2^b-1) \\
   &=\text{clip}(\text{log}_2\lfloor \frac{\sum \text{i\mbox{-}exp}(X_Q)}{ \text{i\mbox{-}exp}(X_Q)} \rceil,0,2^b-1).
   \label{eq:LSI}
   \end{aligned}
\end{equation} 

\subsection{Tailored Row-Stationary Dataflow}
\label{sec:rs_dataflow}

Considering both the inter-layer and intra-layer data dependencies within ViTs' computation and the chunk-based design in our P$^2$-ViT's accelerator, where limited computation resources are partitioned into multiple smaller tailored chunks, we introduce a tailored row-stationary dataflow to size the pipeline processing opportunity introduced by our PoT scaling factors for enhancing throughput (see \emph{Design Consideration \# 2} in Sec. \ref{sec:hardware_considerations} for details).
Here we first take MatMuls of $A \times B = C$ as the example to illustrate the main idea of our row-stationary dataflow: {row} $A_i$ of $A$ is \underline{\emph{spatially}} mapped on the PE array to conduct $A_i \times B$ for generating the corresponding {row} $C_i$ of $C$, and different rows of $A$ are \underline{\emph{temporarily}} processed to sequentially obtain different rows of $C$. 
Next, we will introduce how to tailor the above row-stationary dataflow for our fully quantized ViTs to facilitate both inter-layer and intra-layer pipelines. 

\textbf{Inter-Layer Pipeline.}
To facilitate the inter-layer pipeline, i.e., concurrently executing LN and its subsequent layer on corresponding chunks to enhance hardware utilization and overall throughput, we process LN's input row by row with its tailored LN module for sequentially generating LN's corresponding output rows. Then, they are immediately passed to other chunks for executing the following steps.
Specifically, 
\underline{(i)} as for the computation within the \textbf{LN module}, elements in each row of LN's input are first pre-processed via shifters (see \emph{Step ${1}$} in Fig. \ref{fig:LN}a) and then accumulated to compute mean and variance (see \emph{Step ${2}$} in Fig. \ref{fig:LN}a), which are later utilized to generate elements in the corresponding row of LN's output (see \emph{Step ${3}$}).
Note that besides the optimization for \emph{Step $2$}, where a MAS array is constructed to implement Eq. (\ref{eq:var_2}) for avoiding the latency overhead, the above three steps can also be pipelined to further boost the throughput of our LN module.
Concretely, when consuming \emph{previously} obtained mean and variance to compute \emph{Step $3$}, \emph{Step $1$} paired with \emph{Step $2$} can be concurrently executed to prepare statistics for the \emph{next} row.

\begin{figure}[t]
\centerline{\includegraphics[width=\linewidth]{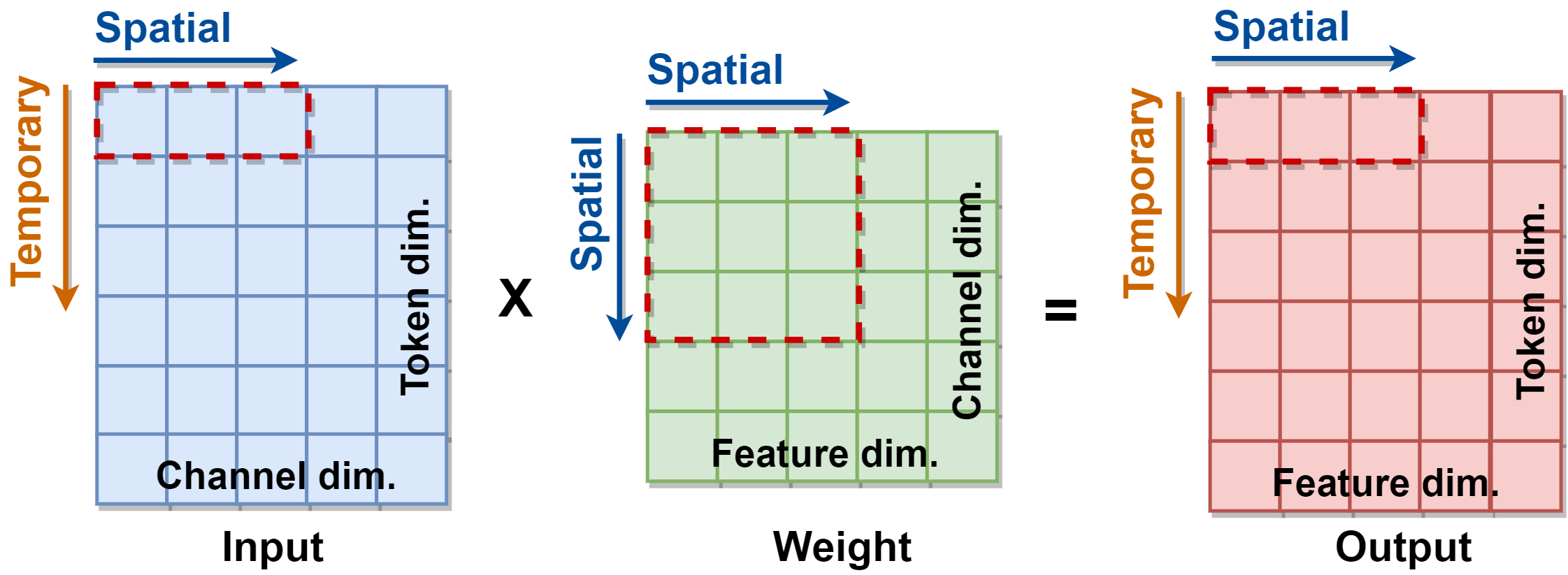}}
	\vspace{-0.6em}
	\caption{The spatial and temporal mappings of our row-stationary dataflow.} 
	\label{fig:rs_dataflow} \vspace{-1.2em}
\end{figure}

\underline{(ii)} After elements in LN's output row are generated by the LN module, they are immediately transferred to the \textbf{re-quantization module} to be re-quantized via shifters.
\underline{(iii)} Then, the re-quantized ones are transmitted to the \textbf{PS-MAC array} to serve as input for computation with LN's subsequent layer.
As illustrated in Fig. \ref{fig:rs_dataflow}, the generated input row is \emph{spatially} mapped to our PS-MAC array to be simultaneously multiplied with multiple columns of the weight, generating the corresponding row of output.
Concurrently, the input's \emph{next} row is prepared by the LN module and re-quantization module.
By doing this, different input rows can be \emph{temporarily} executed on the PS-MAC array to sequentially generate corresponding output rows without delay.
Hence, the timeline of the three modules, i.e., the LN module, re-quantization module, and PS-MAC array, is highly overlapped to enhance throughput. 

\textbf{Intra-Layer Pipeline.}
Here we introduce how to leverage the tailored row-stationary dataflow to achieve ViTs' intra-layer pipeline, i.e., the computation steps in Eq. (\ref{eq:attn}) (including $Q\cdot K^T$, Softmax($\cdot$), and MatMuls with $V$) are concurrently executed on their corresponding chunks for achieving the rapid generation of attention outputs $A$. 
Specifically,
\underline{(i)} each row of $Q$ is \emph{spatially} mapped to our \textbf{PS-MAC array} for being simultaneously multiplied with multiple rows of $K$ to generate the corresponding row of output, 
which is then \underline{(ii)} transferred to our \textbf{Softmax module} for normalization.
\underline{(iii)} After that, the normalized output row is \emph{spatially} mapped to our \textbf{shifter array} to be simultaneously multiplied with multiple columns of $V$ for generating the corresponding row of $A$.
Concurrently, the \emph{next} row of normalized output is prepared by the PS-MAC array and Softmax module.  
Thus, different rows of normalized output can be \emph{temporarily} executed on the shifter array to sequentially and promptly generate the corresponding rows of $A$, thereby boosting both hardware utilization and throughput.

\rb{As for the remaining {computation-intensive linear operations}, i.e., MatMuls in linear projections of MSAs and MatMuls within MLPs, we compute them using our proposed row-stationary dataflow \textbf{\textit{layer by layer}}. Specifically, as illustrated in Fig. \ref{fig:rs_dataflow}, each row of input is spatially mapped to our PS-MAC array for being simultaneously multiplied with multiple rows of weight to generate the corresponding row of output.
Consequently, we only need to cache either a single layer of weights to support layer-by-layer computations or a few adjacent layers to enable pipeline processing on the chip at any given time, thus allowing for the efficient execution of large models under limited buffer constraints.}
\section{Experimental Results}
\label{sec:exp}
\subsection{Experiment Setup}
\textbf{Models, Dataset, and Quantization Details.}
To validate our P$^2$-ViT, 
we consider \underline{\emph{\textbf{four}}} standard {ViTs}: ViT-Base \cite{vit}, and DeiT-Base/Small/Tiny \cite{deit}, and evaluate their quantization performance on {{ImageNet}} \cite{Deng2009ImageNetAL}.
As for \underline{{quantization details,}}
\rall{we randomly select $100$ images from ImageNet's training set as the calibration data, which are used to analyze activation distributions for offline calculating scaling factors of activations and weights \cite{Liu2021PostTrainingQF, Lin2021FQViTPQ, Xiao2022SmoothQuantAA}}, then evaluate accuracy on its validation set.
As for P$^2$-ViT's mixed-precision quantization, weight bit-width choices are set to \{4,8\}, {aiming to minimize the reconfigurable overhead of our precision-scalable PEs and simplify memory management}. For our \underline{evolutionary} algorithm-based fine-grained mixed-precision quantization, the population size is set to $25$, the size and probability of mutation/crossover are set to $10$ and $0.5$, respectively, and we run the evolutionary search for $20$ iterations.

\textbf{Baselines and Evaluation Metrics.}
\underline{(i)} To demonstrate the superiority of P$^2$-ViT's post-training quantization algorithm, we consider \underline{\emph{\textbf{eight}}} baselines: MinMax, EMA \cite{EMA}, Percentile \cite{Percentile}, OMSE \cite{omse}, Bit-Split \cite{bitsplit}, EasyQuant \cite{Wu2020EasyQuantPQ}, PTQ for ViTs \cite{Liu2021PostTrainingQF}, and FQ-ViT \cite{Lin2021FQViTPQ}, and compare with them in top-$1$ accuracy.
{\underline{(ii)} 
To verify P$^2$-ViT's accelerator, we consider \underline{\emph{\textbf{nine}}} SOTA baselines: 
full-precision ViTs executed on two widely-used general computing platforms, including i) Edge GPU (NVIDIA Tegra X2) and ii) GPU (NVIDIA 2080Ti); iii) $8$-bit ViTs quantized via FasterTransformer \cite{fastertransformer}, where non-linear operations remain in ﬂoating-point, and profiled on NVIDIA 2080Ti GPU; $8$-bit ViTs quantized via iv) I-BERT \cite{Kim2021IBERTIB} and v) I-ViT \cite{Li2022IViTIQ}, both of which achieve integer-only inference and are thus implemented and accelerated on Turing Tensor Cores of GPU (NVIDIA 2080Ti) via TVM; and four SOTA quantization-based Transformer accelerators, including vi) Auto-ViT-Acc \cite{Li2022AutoViTAccAF}, vii) FQ-BERT\cite{Liu2021HardwareAO}, viii) HeatViT \cite{dong2023heatvit}, and ix) \cite{huang2023integer}. 

\underline{\textit{\textbf{Note that}}} {\underline{(i)} our algorithm is built upon FQ-ViT \cite{Lin2021FQViTPQ} and the goal is to enhance re-quantization efficiency, making FQ-ViT our most competitive algorithmic baseline.} \underline{(ii)} Quantization methods within the above hardware benchmarks are proposed for QAT while our P$^2$-ViT targets PTQ, thus it is trivial to compare accuracy with them.
However, they serve as competitive hardware benchmarks to compare hardware efﬁciency with our accelerator, as they implement quantitated ViTs on GPU's dedicated integer arithmetic units (Turing Tensor Cores) or tailored accelerators, thus gaining quantization speedups.

\textbf{Hardware Experiment Setup.}
\underline{Characteristics:}
Our P$^2$-ViT's accelerator is designed with a total area of $3.07$mm$^2$ and a total power of $491$mW, and equipped with \underline{(i)} $403$KB global buffers, including $3\times37$KB buffers allocated for queries, keys, and values, $74$KB input buffers, $74$KB output buffers, and $144$KB weight buffers. 
\rb{The buffer sizes are tailored based on the shapes of layers within our executed models, aiming to maximize data reuse opportunities under limited hardware resources.}
\underline{(ii)} As for dedicated chunks, the parallelism of the LN/re-quantization/Softmax module is $64/32/64$. The PS-MAC array consists of $32\times64$ PS-MACs, each of which can conduct one multiplication between $8$-bit activation and $8$-bit weight or two multiplications between $8$-bit activations and $4$-bit weights for supporting proposed mixed-precision quantization. Moreover, the shifter array includes $32\times64$ shifters. The synthesized results indicate that the clock frequency of our P$^2$-ViT's accelerator can be set to $500$MHz, so that we can maximize the computation efficiency without timing violation. 
\underline{Evaluation:} 
We implement a cycle-accurate simulator for P$^2$-ViT's accelerator to obtain fast and reliable estimations, and verify them against the corresponding RTL implementations to ensure the correctness following \cite{Dass2022ViTALiTyUL}.
The adopted unit energy and area are synthesized under a $28$nm CMOS technology using Synopsys tools (e.g., Design Compiler for gate-level netlist \cite{DC}) at a frequency of $500$MHz.
\rall{The energy and latency are measured based on \cite{predictor}, which leverages model structure, hardware architecture, dataflow, and technology-dependent unit
costs (e.g., unit energy/latency costs of computation units and memory accesses across various memory hierarchies) to estimate total energy and latency costs.}

\subsection{Evaluation of P$^2$-ViT's Post-Training Quantization}
\label{sec:alg_results}
\begin{table}[]
\centering
\caption{Top-1 Accuracy comparisons on ImageNet} \vspace{-0.8em}
\setlength{\tabcolsep}{0.5em}
\resizebox{\linewidth}{!}{
\begin{tabular}{l|ccc|cccc} \hline \hline
\multirow{2}{*}{\textbf{Method}}     & \multicolumn{3}{c|}{\textbf{Hardware Efficiency}}    & \multicolumn{4}{c}{\textbf{Accuracy (\%)}} \\ \cline{2-8}
 & \textbf{FQ$^*$}       & \textbf{W/A/Attn}               & \textbf{PoT-S$^\dagger$}                                   & \textbf{DeiT-T}            & \textbf{DeiT-S}           & \textbf{DeiT-B}            & \textbf{ViT-B}             \\ \hline \hline
\rowcolor{darker-blue!50} Full Precision              & {\color{purple}{\xmark}}                    & 32/32/32               & {\color{purple}{\xmark}}                                  & 72.21                & 79.85                & 81.85                & 84.53                \\ \hline
Base PTQ        & {\color{purple}{\xmark}}                    & {8/8/8}                &  {{\color{purple}{\xmark}}}                               & 71.78                & 79.35                & 81.37                & 83.48                \\ \hline
MinMax          & \multirow{4}{*}{\color{dark-green}{\cmark}}   & \multirow{4}{*}{8/8/8} &  \multirow{4}{*}{{\color{purple}{\xmark}}}                                       & 70.94                & 75.05                & 78.02                & 23.64                \\
EMA\cite{EMA}               &                           &                        &                                         & 71.17                & 75.71                & 78.82                & 30.3                 \\
Percentile\cite{Percentile} &                           &                        &                                         & 71.47                & 76.57                & 78.37                & 46.69                \\
OMSE \cite{omse}            &                           &                        &                                         & 71.30                 & 75.03                & 79.57                & 73.39                \\ \hline
Bit-Split \cite{bitsplit}   & \multirow{3}{*}{{\color{purple}{\xmark}}}   & \multirow{3}{*}{8/8/8} &  \multirow{3}{*}{{\color{purple}{\xmark}}}                & -                    & 77.06                & 79.42                & -                    \\
EasyQuant\cite{Wu2020EasyQuantPQ}       &                           &                        &                                         & -                    & 76.59                & 79.36                & -                    \\
PTQ for ViT \cite{Liu2021PostTrainingQF}     &                           &                        &                                         & -                    & 77.47                & 80.48                & -                    \\ \hline
FQ-ViT \cite{Lin2021FQViTPQ}& {{\color{dark-green}{\cmark}}}                  & {8/8/4}                &  {\color{purple}{\xmark}}                & 71.07                & 78.40                 & 80.85                & 82.68                \\
\rowcolor{dark-green!16}\textbf{Ours}            &{\color{dark-green}{\cmark}}                & {8/8/4}                &  \color{dark-green}{\cmark}                           & \textbf{70.92} & \textbf{78.24} & \textbf{80.96} & \textbf{82.80} \\ \hline \hline
Base PTQ        & {\color{purple}{\xmark}}                    & 4/8/8            & {\color{purple}{\xmark}}   & 65.63                & 76.07                & 79.65                & 80.67                \\ \hline
MinMax          & \multirow{4}{*}{\color{dark-green}{\cmark}}   & \multirow{4}{*}{4/8/8}                       & \multirow{4}{*}{{\color{purple}{\xmark}}}                                        & 64.89                & 70.81                & 74.49                & 20.74                \\
EMA \cite{EMA}             &                           &                        &                                         & 65.04                & 71.38                & 75.57                & 29.29                \\
Percentile \cite{Percentile}     &                           &                        &                                         & 65.22                & 72.48                & 76.67                & 29.19                \\
OMSE \cite{omse}           &                           &                        &                                         & 65.13                & 70.01                & 77.85                & 63.78                \\ \hline
FQ-ViT \cite{Lin2021FQViTPQ}         & {\color{dark-green}{\cmark}}                  & 4/8/4                  & \color{purple}{\xmark}                                        & 64.86                & 75.34                & 79.00                & 80.67                \\ 
\rowcolor{dark-green!16}\textbf{Ours}            &{\color{dark-green}{\cmark}}                & 4/8/4                        & \color{dark-green}{\cmark}                   & \textbf{65.64}       & \textbf{75.29}       & \textbf{79.47}       & \textbf{80.20}  \\ \hline \hline
\rowcolor{dark-green!16}\textbf{Ours}            &{\color{dark-green}{\cmark}}                & M$^\diamond$/8/4                        & \color{dark-green}{\cmark}                   & \textbf{67.96}           & \textbf{76.29}           & \textbf{80.13}          & \textbf{81.14}  \\ \hline 
\rowcolor{dark-green!16}\textbf{Ours}            &{\color{dark-green}{\cmark}}                & M$^\triangleleft$/8/4                        & \color{dark-green}{\cmark}                   & \textbf{69.60}           & \textbf{77.56}           & \textbf{80.59}          & \textbf{82.10}   \\ \hline \hline
\end{tabular}} \label{tab:acc_compare} 
\begin{tablenotes}
		\footnotesize
		\item[$\ast$] $\ast$ is the abbreviation of fully quantized, and indicates that all LN and Softmax in ViTs are quantized; $\dagger$ denotes PoT scaling factors.
        \item[$\ast$] $\diamond$ represents the mixed-precision between $4$-bit and $8$-bit for weights with $\mathbf{1.1\times}$ model size overhead over the W$4$ counterparts, while $\triangleleft$ denotes the mixed-precision quantization with $\mathbf{1.5\times}$ model size overhead.
	  \end{tablenotes} \vspace{-1.4em}
\end{table}
\textbf{Comparisons with SOTA Baselines.}
As shown in Table \ref{tab:acc_compare}, our P$^2$-ViT's post-training quantization (PTQ) consistently surpasses all SOTA baselines regarding trade-offs between quantization accuracy and hardware efficiency (e.g., the enhanced re-quantization efficiency by PoT scaling factors), verifying our effectiveness.
Specifically, 
\underline{(i)} as for comparisons with the base PTQ, \rc{which is the most straightforward method that adopts MinMax to quantize MatMuls in ViTs, while leaving scaling factors, LN, and Softmax in floating-point}, we can gain nearly lossless quantization performance ({i.e., $\downarrow$$0.41$\%$\sim${$\downarrow$$1.12$\%} accuracy when both weights and activations in ViTs are represented in $8$-bit, and $\downarrow$$0.78$\%$\sim$$\uparrow$$0.01$\% accuracy when weights are compressed to $4$-bit}) while offering more hardware-friendly ViTs. 
\rall{Particularly, we i) further quantize the most sensitive non-linear LN and Softmax in ViTs to offer fully quantized ViTs, and ii) convert floating-point scaling factors into the PoT format for minimizing re-quantization overhead.}
\underline{(ii)} \rb{When compared with FQ-ViT \cite{Lin2021FQViTPQ}, our counterpart with floating-point scaling factors and the most competitive baseline, we can achieve comparable quantization performance while significantly boosting hardware efficiency.
Particularly, we gain {$\downarrow$${0.16}$\%$\sim$$\uparrow$${0.12}$ accuracy when the bit-width of weights/activations/attention maps (W/A/Attn) in ViTs is $8/8/4$, and $\downarrow$${0.47}$\%$\sim$$\uparrow$$\mathbf{0.78}$ accuracy with even lower bit-width for W/A/Attn, which is $4/8/4$}.
Meanwhile, we boost the hardware efficiency of fully quantized ViTs from \emph{\textbf{two perspectives}}: i) hardware-friendly symmetric quantization for activations and ii) PoT scaling factors. }

\underline{(iii)} Despite the hardware efficiency of ViTs with W$4$, there exist non-negligible accuracy drops when compared with their W$8$ counterparts.  
As a solution, our P$^2$-ViT is further equipped with coarse-to-fine automatic mixed-precision quantization and offers {$\uparrow$$\mathbf{0.66}$\%$\sim$$\uparrow$$\mathbf{2.32}$\% and $\uparrow$$\mathbf{1.12}$\%$\sim$$\uparrow$$\mathbf{3.96}$\% accuracy with only $\mathbf{1.1}\times$ and $\mathbf{1.5}\times$ model size overheads, respectively, when compared with the W4 ones.}

\begin{table}[]
\centering
\caption{Ablation studies of P$^2$-ViT's dedicated quantization scheme} \vspace{-0.8em}
\setlength{\tabcolsep}{0.5em}
\resizebox{0.92\linewidth}{!}{
\begin{tabular}{ccc|cccc}
\hline \hline \color{black}{\ding{182}}          & \textbf{\color{black}{\ding{183}}}          & \textbf{W/A/Attn}      & \textbf{DeiT-Tiny} & \textbf{DeiT-Small} & \textbf{DeiT-Base} & \textbf{ViT-Base} \\ \hline \hline
{\color{purple}{\xmark}} & {\color{purple}{\xmark}} & \multirow{4}{*}{8/8/4} & 69.80              & 76.15               & 80.27              & 80.36           \\
\color{dark-green}{\cmark} & {\color{purple}{\xmark}} &                        & 70.82              & 77.26               & 80.61              & 82.27             \\
{\color{purple}{\xmark}} & \color{dark-green}{\cmark} &                        & 69.93              & 77.76               & 80.79              & 80.68     \\             
\rowcolor{dark-green!16} \color{dark-green}{\cmark} & \color{dark-green}{\cmark} &                        & \textbf{70.92}       & \textbf{78.24}    & \textbf{80.96}                    & \textbf{82.80}                   \\ \hline \hline
 {\color{purple}{\xmark}} & {\color{purple}{\xmark}} & \multirow{4}{*}{4/8/4} & 63.39              & 71.94               & 78.01              & 75.65                   \\
\color{dark-green}{\cmark} & {\color{purple}{\xmark}} &                        & 65.54              & 73.88               & 79.11              & 78.62             \\
{\color{purple}{\xmark}} & \color{dark-green}{\cmark} &                        & 63.77              & 72.93               & 78.76                    & 75.74                   \\  
\rowcolor{dark-green!16} \color{dark-green}{\cmark} & \color{dark-green}{\cmark} &                        & \textbf{65.64}       & \textbf{75.29}       & \textbf{79.47}              & \textbf{80.20}  \\ \hline \hline          
\end{tabular}} 
\label{tab:ablation_quantization_scheme}
\begin{tablenotes}
		\footnotesize
		\item[$\ast$] \color{black}{\ding{182}} represents adaptive PoT rounding; 
        \color{black}{\ding{183}} denotes PoT-aware smoothing.
	  \end{tablenotes} \vspace{-0.8em}
\end{table} 
\textbf{{Effectiveness of the Dedicated Quantization Scheme.}}
As shown in Table \ref{tab:ablation_quantization_scheme}, we see that \underline{(i)} directly rounding floating-point scaling factors in fully quantized ViTs to their nearest PoT ones yields severe accuracy drops, calling for our dedicated quantization scheme to enhance accuracy.  
Specifically, \underline{(ii)} our adaptive PoT rounding offers $\uparrow$$\mathbf{0.35}$\%$\sim$$\uparrow$$\mathbf{1.91}$\% and $\uparrow$$\mathbf{1.10}$\%$\sim$$\uparrow$$\mathbf{2.97}$\% accuracy in ViTs with W$8$ and W$4$, respectively.
\underline{(iii)} Besides, our PoT-Aware smoothing offers $\uparrow$$\mathbf{0.13}$\%$\sim$$\uparrow$$\mathbf{1.61}$\% and $\uparrow$$\mathbf{0.09}$\%$\sim$$\uparrow$$\mathbf{1.00}$\% accuracy in W$8$ and W$4$ ViTs, respectively.
\underline{(iv)} By integrating these two enablers, we achieve the best accuracy, i.e., {$\uparrow$$\mathbf{0.69}$\%$\sim$$\uparrow$$\mathbf{2.44}$\% and $\uparrow$$\mathbf{1.46}$\%$\sim$$\uparrow$$\mathbf{4.56}$\% accuracy in W$8$ and W$4$ ViTs, respectively}. 

\begin{table}[]
\setlength{\tabcolsep}{0.33em}
\centering
\caption{Quantization accuracy under W6A6 when adopting our method upon RepQ-ViT \cite{li2023repq}. ``PoT-S" donates PoT scaling factors and ``RTN" means rounding-to-nearest} \vspace{-0.5em}
\resizebox{\linewidth}{!}{
\begin{tabular}{c|c|ccccc} \hline \hline
\textbf{Methods} & \textbf{PoT-S}        & \textbf{DeiT-Tiny} & \textbf{DeiT-Small} & \textbf{DeiT-Base} & \textbf{ViT-Small} & \textbf{ViT-Base} \\ \hline \hline
RepQ-ViT \cite{li2023repq}        & \color{purple}{\xmark} & 70.76              & 78.90               & 81.27              & 80.43              & 83.62             \\
RTN              & \color{dark-green}{\cmark} & 66.04              & 75.11               & 80.37              & 77.68              & 81.62             \\
\rowcolor{dark-green!16} \textbf{Ours}    & \color{dark-green}{\cmark} & \textbf{70.55}     & \textbf{78.59}      & \textbf{81.03}     & \textbf{79.66}     & \textbf{83.58} \\ \hline \hline  
\end{tabular}} \label{tab:scalibility} \vspace{-1.6em}
\end{table}

\rc{\textbf{{Scalability of the Dedicated Quantization Scheme.}} To validate the scalability of our dedicated quantization scheme, we further verify it on top of another PTQ framework, RepQ-ViT \cite{li2023repq}.
Specifically, RepQ-ViT focuses on developing a dedicated quantization method, i.e., quantization scale reparameterization strategy, to efficiently and effectively quantize sensitive activations and thus boost \textbf{\textit{quantization accuracy}}. In contrast, our work targets converting floating-point scaling factors to PoT ones, thus minimizing re-quantization overhead and enhancing \textbf{\textit{hardware efficiency}}.
As we focus on complementary aspects, our approach can be combined with RepQ-ViT to marry the best of both worlds as well as verifying our scalability. As listed in Table \ref{tab:scalibility}, \underline{(i)} when implementing our method on top of RepQ-ViT \cite{li2023repq}, we can offer PoT scaling factors with negligible accuracy drops ($0.04\%$$\sim$$0.77\%$), demonstrating our scalability.
\underline{(ii)} We achieve much higher accuracy compared to the widely adopted rounding-to-nearest (RTN) method, which yields $0.90\%$$\sim$$4.72\%$ accuracy degradation, further verifying our effectiveness.
}

\begin{table}[]
\centering
\caption{Ablation studies of P$^2$-ViT's coarse-to-fine mixed-precision quantization. ``MS" donates model size (MB)} \vspace{-0.8em}
\setlength{\tabcolsep}{0.5em}
\resizebox{\linewidth}{!}{
\begin{tabular}{ccc|cc|cc|cc|cc}
\hline \hline 
\multirow{2}{*}{\textbf{Hessian}} & \multicolumn{1}{c}{\multirow{2}{*}{\textbf{Evo}}}  & \multirow{2}{*}{\textbf{W/A/Attn}} & \multicolumn{2}{c|}{\textbf{DeiT-Tiny}}                     & \multicolumn{2}{c|}{\textbf{DeiT-Small}} & \multicolumn{2}{c|}{\textbf{DeiT-Base}} & \multicolumn{2}{c}{\textbf{ViT-Base}}   \\ \cline{4-11} 
                         & \multicolumn{1}{c}{}                      &                           & \textbf{Acc.}           & \multicolumn{1}{c|}{\textbf{MS}}           & \textbf{Acc.}            & \textbf{MS}            & \textbf{Acc.}           & \textbf{MS}            & \textbf{Acc.}           & \textbf{MS}            \\ \hline \hline 
\rowcolor{darker-blue!50}\color{purple}{\xmark}    & \color{purple}{\xmark}                     & 4/8/4                     & 65.64          & \multicolumn{1}{c|}{2.5}          & 75.29           & 11.0          & 79.47          & 43.0          & 80.20          & 43.0          \\
\rowcolor{darker-blue!50}\color{purple}{\xmark}    & \color{purple}{\xmark}                     & 8/8/4                     & 70.92          & \multicolumn{1}{c|}{5.0}          & 78.24           & 22.0          & 80.96          & 86.0          & 82.80          & 86.0          \\ \hline
\color{dark-green}{\cmark}    & \color{purple}{\xmark}                     & M/8/4                     & 67.76          & \multicolumn{1}{c|}{2.8}          & 75.80           & 12.1          & 79.85          & 47.3          & 80.94          & 47.3          \\
\rowcolor{dark-green!16} \color{dark-green}{\cmark}    & \color{dark-green}{\cmark}                     & M/8/4                     & \textbf{67.96} & \textbf{2.8}                      & \textbf{76.29}  & \textbf{12.1} & \textbf{80.13} & \textbf{47.3} & \textbf{81.14} & \textbf{47.3} \\ \hline
\color{dark-green}{\cmark}    & \color{purple}{\xmark}                     & M/8/4                     & 69.22          & \multicolumn{1}{c|}{3.8}          & 76.80           & 16.5          & 80.38          & 64.5          & 81.71          & 64.5          \\
\rowcolor{dark-green!16} \color{dark-green}{\cmark}    & {\color{dark-green}{\cmark}} & M/8/4                     & \textbf{69.60} & \multicolumn{1}{c|}{\textbf{3.8}} & \textbf{77.56}  & \textbf{16.5} & \textbf{80.59} & \textbf{64.5} & \textbf{82.10} & \textbf{64.5} \\ \hline \hline 
\end{tabular}} \label{tab:ablation_mixed} \vspace{-0.6em}
\end{table}

\rb{\textbf{{Effectiveness of Our Coarse-to-Fine Automatic Mixed-Precision Quantization.}}
As listed in Table \ref{tab:ablation_mixed}, our mixed-precision quantization consistently outperforms fixed-precision quantization.
Specifically, 
\underline{(i)} when only involving Hessian-based mixed-precision quantization, we offer {$\uparrow$$\mathbf{0.38}$\%$\sim$$\uparrow$$\mathbf{2.12}$\% and $\uparrow$$\mathbf{0.90}$\%$\sim$$\uparrow$$\mathbf{3.58}$\%} accuracy with $1.1\times$ and $1.5\times$ model size overheads, respectively, compared to counterparts with fixed 4-bit weights.
Furthermore, \underline{(ii)} when further integrating with the evolutionary search algorithm, we can correspondingly achieve {$\uparrow$$\mathbf{0.66}$\%$\sim$$\uparrow$$\mathbf{2.32}$\% and $\uparrow$$\mathbf{1.12}$\%$\sim$$\uparrow$$\mathbf{3.96}$\%} accuracy, demonstrating the significance of our P$^2$-ViT's coarse-to-fine automatic mixed-precision quantization, which integrates both Hessian-based coarse-grained and evolutionary algorithm-based fine-grained enablers to offer better trade-offs between search efficiency and accuracy.}

\vspace{-0.5em}
\subsection{Evaluation of P$^2$-ViT's Accelerator}
\begin{figure}[t]
	\centerline{\includegraphics[width=\linewidth]{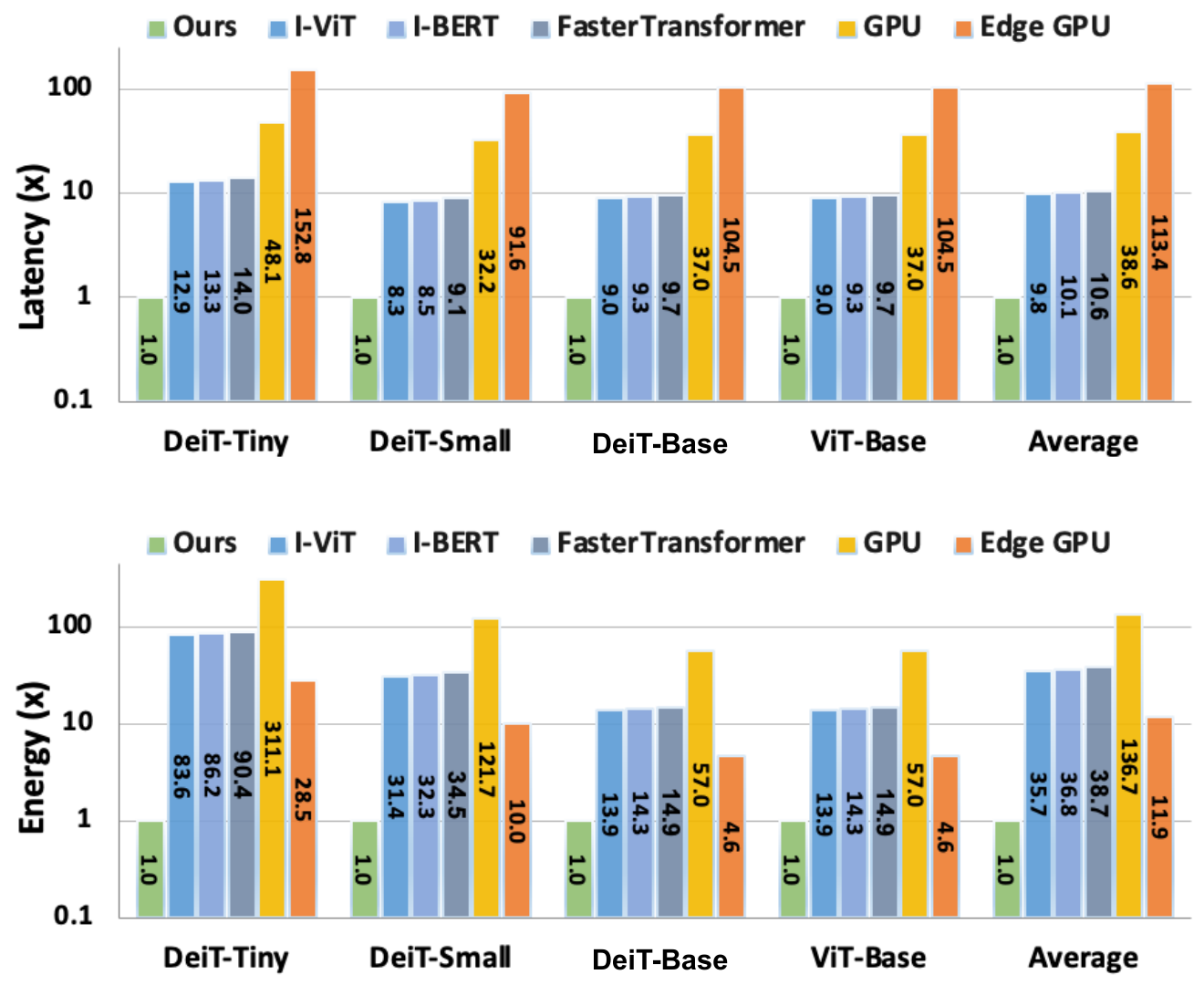}}
	\vspace{-0.6em}
	\caption{The latency and energy comparisons of our P$^2$-ViT's accelerator for implementing P$^2$-ViT's offered $8$-bit ViTs against several SOTA GPU baselines. \textbf{Note} that the y-axis is expressed in logarithms for better illustration. \rb{Besides, FastTransformer \cite{fastertransformer}, I-BERT \cite{Kim2021IBERTIB}, and I-ViT \cite{Li2022IViTIQ} mentioned here are not transformer models themselves but {{quantization acceleration strategies}}, which can be implemented on GPUs to facilitate actual hardware speedup.}} 
	\label{fig:hardware_results} \vspace{-0.9em}
\end{figure}
\textbf{Comparisons with SOTA Baselines on GPUs.}
For fair comparisons, we follow \cite{Qu2022DOTADA,Ham2021ELSAHC} to scale up the hardware resources of our accelerator to have comparable peak throughput of GPUs. \rb{As shown in Fig. \ref{fig:hardware_results}, our P$^2$-ViT's accelerator consistently outperforms all SOTA baselines, including the GPU equipped with {quantization acceleration strategies} (such as FastTransformer \cite{fastertransformer}, I-BERT \cite{Kim2021IBERTIB}, and I-ViT \cite{Li2022IViTIQ}), the vanilla edge GPU, and vanilla GPU.} 
Specifically, 
\underline{(i)} when compared with the full-precision ViTs on GPU and Edge GPU, we can offer average $\mathbf{38.6\times}$ and $\mathbf{113.4\times}$ speedups, and $\mathbf{136.7\times}$ and $\mathbf{11.9\times}$ energy savings, respectively, when executing our $8$-bit fully quantized ViTs on our dedicated accelerator.
\underline{(ii)} When compared with the most competitive baselines, i.e., the 8-bit ViTs quantized by I-ViT \cite{Li2022IViTIQ} and I-BERT \cite{Kim2021IBERTIB}, and accelerated on GPU's dedicated integer arithmetic units (Turing Tensor Cores), we can achieve average $\mathbf{9.8\times}$ and $\mathbf{10.1\times}$ speedups, and $\mathbf{35.7\times}$ and $\mathbf{36.8\times}$ energy savings, respectively.
We attribute latency benefits to P$^2$-ViT's tailored row-stationary dataflow, which facilitates both inter-layer and intra-layer pipelines when executing fully quantized ViTs on our chunk-based design.
Energy improvements result mainly from our re-quantization module that adopts shifters to re-quantize activations with PoT scaling factors for minimizing the re-quantization overhead.


\begin{table}[t]
\centering
\caption{Comparisons with the SOTA ViT accelerator Auto-ViT-Acc \cite{Li2022AutoViTAccAF}} \vspace{-0.8em}
\setlength{\tabcolsep}{0.3em}
\resizebox{\linewidth}{!}{
\begin{tabular}{c|c|cc|cc|c}
\hline \hline
\multirow{2}{*}{\textbf{Methods}}                                        & \textbf{Formats} & \multicolumn{2}{c|}{\textbf{DeiT-Small}} & \multicolumn{2}{c|}{\textbf{DeiT-Base}} & \textbf{Average} \\ \cline{2-6}
                                                                         & \textbf{W8A8}    & \textbf{FPS/DSP}   & \textbf{GOPS/DSP}   & \textbf{FPS/DSP}   & \textbf{GOPS/DSP}  & \textbf{Imprv. ($\times$)}   \\ \hline \hline
\multirow{2}{*}{\begin{tabular}[c]{@{}c@{}}Auto-ViT-\\Acc \cite{Li2022AutoViTAccAF}\end{tabular}} & Fixed            & 0.040              & 0.367               & 0.013              & 0.435              & \textbf{1.00}    \\
                                                                         & Fixed+PoT        & 0.064              & 0.585               & 0.022              & 0.759              & \textbf{1.74}    \\ \hline
\rowcolor{dark-green!16} \textbf{Ours}                                                            & \textbf{Fixed}            & \textbf{0.068}     & \textbf{0.623}      & \textbf{0.023}     & \textbf{0.799}     & \textbf{1.84}    \\ \hline \hline
\end{tabular}} \label{tab:compare-to-auto-vit-acc} \vspace{-1em}
\end{table}

\begin{table}[t]
\centering
\caption{Comparisons to SOTA Transformer accelerator FQ-BERT \cite{Liu2021HardwareAO}} \vspace{-0.8em}
\setlength{\tabcolsep}{0.2em}
\resizebox{\linewidth}{!}{
\begin{tabular}{c|c|cccc|c}
\hline \hline
\textbf{Methods}  & \textbf{FQ-BERT} \cite{Liu2021HardwareAO} & \multicolumn{4}{c|}{\textbf{\textcolor{dark-green}{Ours}}}            & \textbf{Average} \\ \cline{1-6}
\textbf{Models}   & \textbf{BERT} \cite{bert}    & \textbf{DeiT-Tiny} & \textbf{DeiT-Small} & \textbf{DeiT-Base} & \textbf{ViT-Base} & \textbf{Imprv}. ($\times$)   \\ \hline \hline
\textbf{GOPS/DSP} & 0.79    & \textbf{1.64}      & \textbf{1.70}       & \textbf{1.69}      & \textbf{1.69 }    & \textbf{2.13}    \\ \hline \hline
\end{tabular}} \label{tab:compare-to-fqbert} \vspace{-1.3em}
\end{table}

\begin{table}[t]
\centering
\caption{Comparisons to SOTA ViT accelerators \cite{dong2023heatvit, huang2023integer}} \vspace{-0.8em}
\setlength{\tabcolsep}{0.2em}
\resizebox{\linewidth}{!}{
\begin{tabular}{c|ccc|ccc}
\hline \hline
\textbf{Models}   & \multicolumn{3}{c|}{\textbf{DeiT-Tiny}} & \multicolumn{3}{c}{\textbf{DeiT-Small}} \\ \hline 
\textbf{Methods}  & \textbf{HeatViT} \cite{dong2023heatvit}  & \textbf{TCAS-1'23} \cite{huang2023integer}   & \textbf{\textcolor{dark-green}{Ours}}   & \textbf{HeatViT} \cite{dong2023heatvit}  & \textbf{TCAS-1'23} \cite{huang2023integer}  & \textbf{\textcolor{dark-green}{Ours}}   \\ \hline
\textbf{Formats}  & W8       & W8A8       & \textbf{W8A8}   & W8       & W8A8       & \textbf{W8A8}   \\ \hline
\textbf{GOPS/DSP} & 0.160    & 0.486      & \textbf{0.890}  & 0.174    & 0.601      & \textbf{0.923}  \\ \hline \hline
\end{tabular}} \label{tab:exp_hw_vits} \vspace{-1em}
\end{table}
\textbf{Comparisons with SOTA Transformer Accelerators.}
As accelerators dedicated to Transformer quantization are mainly based on FPGAs \cite{Li2022AutoViTAccAF, Liu2021HardwareAO, dong2023heatvit, huang2023integer}, to enable fair comparisons, we implement P$^2$-ViT's accelerator on the same FPGA platform and clock frequency and compare with them in terms of computation efﬁciency (i.e., FPS/DSP and GOPS/DSP). We use widely adopted DSP packing strategies \cite{Xilinx-8bit, Xilinx-4bit} following \cite{Li2022AutoViTAccAF} to pack multiple low-bit multipliers into one DSP to enhance DPS utilization.
Specifically, \rb{\underline{(i)} when executing DeiTs \cite{deit} with fixed W8A8 quantization, we can achieve an average $\uparrow$$\mathbf{1.84\times}$ computation utilization efficiency over the SOTA ViT accelerator Auto-ViT-Acc \cite{Li2022AutoViTAccAF} (see Table \ref{tab:compare-to-auto-vit-acc}). 
\underline{(ii)} In comparison to the SOTA Transformer accelerator FQ-BERT \cite{Liu2021HardwareAO}, we can gain an average $\uparrow$$\mathbf{2.13\times}$ GOPS/DSP (see Table \ref{tab:compare-to-fqbert}).
Besides, \underline{(iii)} when compared with the SOTA ViT accelerator \cite{huang2023integer}, we can gain an average $\uparrow$$\mathbf{1.53\times}$ GOPS/DSP (see Table \ref{tab:exp_hw_vits}).}
The improvements are attributed to our effective quantization algorithm, which trades partial multiplications with bitwise shifts, thus saving the DSP consumption.
Besides, our tailored row-stationary dataflow, which sizes the pipeline opportunity introduced by our PoT scaling factor, also contributes to enhanced throughput.

\begin{table}[t]
\centering
\caption{Ablation studies of our low-bit and mixed-precision quantization when executed on our dedicated accelerator} \vspace{-0.8em}
\setlength{\tabcolsep}{0.4em}
\resizebox{0.96\linewidth}{!}{
\begin{tabular}{c|cccc|cccc} \hline \hline 
\multirow{2}{*}{\textbf{W/A/Attn}} & \multicolumn{4}{c|}{\textbf{Speedup ($\times$)}}                                 & \multicolumn{4}{c}{\textbf{Energy Saving ($\times$)}}                       \\ \cline{2-9}
                                   & \textbf{DeiT-T} & \textbf{DeiT-S} & \textbf{DeiT-B} & \textbf{ViT-B} & \textbf{DeiT-T} & \textbf{DeiT-S} & \textbf{DeiT-B} & \textbf{ViT-B} \\ \hline \hline 
\rowcolor{darker-blue!50} \textbf{8/8/4}                     & 1.00            & 1.00            & 1.00            & 1.00           & 1.00            & 1.00            & 1.00            & 1.00           \\
\textbf{4/8/4}                     & 1.84            & 1.91            & 1.95            & 1.95           & 1.72            & 1.83            & 1.90            & 1.90           \\
\textbf{M$^\diamond$/8/4}                     & 1.75            & 1.78            & 1.86            & 1.82           & 1.65            & 1.71            & 1.81            & 1.78           \\
\textbf{M$^\triangleleft$/8/4}                     & 1.31            & 1.32            & 1.33            & 1.34           & 1.28            & 1.30            & 1.32            & 1.32          \\ \hline  \hline 
\end{tabular}} \label{tab:mixed_hw}
\begin{tablenotes}
		\footnotesize
        \item[$\ast$] $\diamond$ and $\triangleleft$ correspond to the ones with $\mathbf{1.1\times}$ and $\mathbf{1.5\times}$ overheads of model size in Table \ref{tab:ablation_mixed}, respectively.
	  \end{tablenotes} \vspace{-1em}
\end{table}

\textbf{Effectiveness of our Low-Bit Quantization and Mixed-Precision Quantization.}
When adopting our dedicated accelerator to accelerate our fully quantized ViTs with different bit-widths of weights, as listed in Table \ref{tab:mixed_hw}, \underline{(i)} the $4$-bit quantized ones gain $\mathbf{1.84\times}$$\sim$$\mathbf{1.95\times}$ speedups and $\mathbf{1.72\times}$$\sim$$\mathbf{1.90\times}$ energy savings over the $8$-bit counterparts. 
Despite their promising hardware efficiency, they yield $\downarrow$$\mathbf{1.49\%}$$\sim$$\downarrow$$\mathbf{5.28\%}$ accuracy compared to W$8$ ViTs (see Table \ref{tab:ablation_mixed}), calling for mixed-precision quantization.
\underline{(ii)} Particularly, our mixed-precision quantization offers $\mathbf{1.75\times}$$\sim$$\mathbf{1.86\times}$ speedups and $\mathbf{1.65\times}$$\sim$$\mathbf{1.81\times}$ energy savings with $\downarrow$$\mathbf{0.83\%}$$\sim$\textbf{}$\mathbf{2.96\%}$ accuracy over the $8$-bit counterparts.
\underline{(iii)} We can further achieve $\mathbf{1.31\times}$$\sim$$\mathbf{1.34\times}$ speedups and $\mathbf{1.28\times}$$\sim$$\mathbf{1.32\times}$ energy savings with only  $\downarrow$$\mathbf{0.37\%}$$\sim$$\downarrow$$\mathbf{1.32\%}$ accuracy, verifying the effectiveness and flexibility of our mixed-precision quantization for different applications and scenarios, each of which requires different quantization performance and efficiency trade-offs.

\begin{table}[t]
\centering
\caption{Ablation studies of our tailored row-stationary dataflow on the fully quantized DeiT-Tiny \cite{deit} with $8$-bit weights}
\vspace{-0.8em}
\setlength{\tabcolsep}{0.7em}
\begin{tabular}{cc|ccc}
\hline \hline
\multicolumn{2}{c|}{\textbf{Pipeline}}        & \multicolumn{3}{c}{\textbf{Latency ($\times$)}}          \\ \hline
\textbf{Inter-Layer} & \textbf{Intra-layer} & \textbf{Attention} & \textbf{MLP}  & \textbf{Overall} \\ \hline \hline
\rowcolor{darker-blue!50} \color{purple}{\xmark}                 & \color{purple}{\xmark}                & 1.79               & 1.15          & 1.57             \\
\color{dark-green}{\cmark}                 & \color{purple}{\xmark}                & 1.45               & 1.00          & 1.32             \\
\color{purple}{\xmark}                 & \color{dark-green}{\cmark}                & 1.34               & 1.15          & 1.37             \\
\rowcolor{dark-green!16} \color{dark-green}{\cmark}        & \color{dark-green}{\cmark}       & \textbf{1.00}      & \textbf{1.00} & \textbf{1.00}    \\ \hline \hline
\end{tabular} \label{tab:ablation_rs} \vspace{-1.2em}
\end{table}
\textbf{Effectiveness of Our Tailored Row-Stationary Dataflow.}
As shown in Table \ref{tab:ablation_rs}, where we take fully quantized DeiT-Tiny \cite{deit} with $8$-bit weights as the example, our tailored row-stationary dataflow consistently outperforms all baselines in terms of latency, verifying its effectiveness in facilitating both inter-layer (i.e., LN and its subsequent layer) and intra-layer (i.e., the computation within self-attention) pipelines to enhance overall throughput.
Specifically, 
\underline{(i)} as for the acceleration for self-attention blocks, where exists opportunities for both inter-layer and intra-layer pipelines, our tailored row-stationary dataflow can offer up to $\mathbf{1.79\times}$ speedup over the naive one (the row highlighted in gray).
\underline{(ii)} While as for the acceleration for MLP blocks, where only exists the opportunity for the inter-layer pipeline, we achieve $\mathbf{1.15\times}$ speedup over the vanilla one.
\underline{(iii)} Overall, we can gain $\mathbf{1.57\times}$ end-to-end speedup when considering both attention and MLP blocks.

\underline{\textbf{\textit{Note that}}} this superiority also emphasizes our motivation of exploring PoT scaling factors, which enable us to conduct re-quantization on the chip, thus enabling pipeline processing.

\section{Conclusion}
\label{sec:conclusion}
In this paper, we have proposed, developed, and validated {P$^2$-ViT}, a \underline{\textbf{P}}ower-of-Two (PoT) \underline{\textbf{p}}ost-training quantization and acceleration framework to accelerate Vision Transformers (ViTs).
Specifically, we develop {a dedicated quantization scheme} to effectively quantize ViTs with PoT scaling factors.
On top of that, we propose {coarse-to-fine automatic mixed-precision quantization} to achieve better trade-offs between quantization accuracy and hardware efficiency. 
At the hardware level, we develop {a chunk-based accelerator} to alleviate the reconfigurable overhead, and further design {a tailored row-stationary dataflow} to embrace the pipeline opportunity introduced by our PoT scaling factors to promote throughput.
Extensive experiments consistently validate our effectiveness. {Particularity, we gain comparable or even higher quantization performance with PoT scaling factors when compared with the counterpart with floating-point scaling factors, while offering up to $10.1\times$ speedup and $36.8\times$ energy saving over GPU's Turing Tensor Cores and up to $\uparrow$${1.84\times}$ computation utilization efficiency against the SOTA ViT accelerators.}

\ra{\textbf{Potential Impact.} 
Our P$^2$-ViT is expected to shed light on the quantization and acceleration for Transformer-based models, aiming to facilitate their real-world deployment and applications.
Particularly, considering the significant demand for Transformer-based models in real-world scenarios, especially the extensive use of large language models (LLMs) like ChatGPT\footnote{\url{https://openai.com/chatgpt}} and large vision models (LVMs) like DALL-E2\footnote{\url{https://openai.com/dall-e-3}}, coupled with their unaffordable training/fine-tuning costs, we
can expect that our post-training method is potentially well-positioned to compress and accelerate them without necessitating fine-tuning, thus enhancing the immersive user experiences in our daily life.}

\ra{\textbf{{Future Works.}} 
Despite P$^2$-ViT is effective for standard ViTs \cite{deit, vit}, it overlooks the inherent advantages of efficient ViTs \cite{Cai2022EfficientViTEL, Han2023FLattenTV}, where the quadratic computational complexity of vanilla Softmax-based attention is replaced by more efficient Softmax-free attentions with linear computational complexity. Thus, our future research will focus on quantization and acceleration for efficient ViTs to fully unlock their algorithmic benefits.
Furthermore, a more in-depth investigation into the quantization and acceleration dedicated to LLMs and LVMs is highly encouraged to advance their practical implementations and applications.}

\bibliographystyle{IEEEtran}
\bibliography{main}

\end{document}